\documentclass[journal]{IEEEtran}
\usepackage{amsmath,amsfonts}
\usepackage{algorithmic}
\usepackage{algorithm}
\usepackage{array}
\usepackage[caption=false,font=normalsize,labelfont=sf,textfont=sf]{subfig} 
\usepackage{textcomp}
\usepackage{stfloats}
\usepackage{url}
\usepackage{verbatim}
\usepackage{graphicx}
\usepackage{cite}

\usepackage{tikz}
\usepackage{comment}
\usepackage{amsmath,amssymb} 
\usepackage{color}
\usepackage{booktabs}
\usepackage{multirow}
\usepackage{bm}
\usepackage{float}
\usepackage{arydshln} 
\usepackage{setspace}
\usepackage{color}

\newcommand{\tabincell}[2]{\begin{tabular}{@{}#1@{}}#2\end{tabular}}
\usepackage{balance}

\begin{document}
\title{Compact Model Training by Low-Rank Projection with Energy Transfer}
\author{Kailing Guo*,~\IEEEmembership{Member,~IEEE}, Zhenquan Lin*, Canyang Chen, Xiaofen Xing,~\IEEEmembership{Member,~IEEE}, \\ Fang Liu,~\IEEEmembership{Member,~IEEE}, Xiangmin Xu,~\IEEEmembership{Senior Member,~IEEE}

\thanks{Manuscript received 31 December 2022; revised 7 September
2023 and 3 February 2024; accepted 1 May 2024. This work was supported
in part by the Key Research and Development Program of Guangzhou,
China, under Grant 202103020003; in part by the Fundamental Research
Funds for Central Universities, South China University of Technology
(SCUT), under Grant 2023ZYGXZR086 and Grant 2023ZYGXZR013; in
part by the National Key Research and Development Program of China
under Grant 2022YFB4500600; in part by the National Natural Science
Foundation of China under Grant 62102102 and Grant 61802131; in part
by the Basic and Applied Basic Research Foundation of Guangzhou under
Grant 202201010681; in part by the Science and Technology Project of
Guangdong Province under Grant 2022B0101010003; in part by Guangzhou
Key Laboratory of Body Data Science under Grant 201605030011; in part by
the Science and Technology Project of Zhongshan under Grant 2019AG024;
and in part by Guangdong Provincial Key Laboratory of Human Digital Twin
under Grant 2022B1212010004. (Kailing Guo and Zhenquan Lin contributed
equally to this work.) (Corresponding author: Xiangmin Xu.)

Kailing Guo is with the School of Electronic and Information Engineering, South China University of Technology, Guangzhou 510640, China,
and also with the Pazhou Laboratory, Guangzhou 510330, China (e-mail:
guokl@scut.edu.cn).

Fang Liu is with the School of Internet finance and Information Engineering, Guangdong University of Finance, Guangzhou 510521, China (e-mail:
47-032@gduf.edu.cn).

Xiangmin Xu is with the School of Future Technology, and School of Electronic and Information Engineering, South China University of Technology,
Guangzhou 510640, China, also with the Pazhou Laboratory, Guangzhou
510330, China, and also with the Institute of Artificial Intelligence, Hefei
Comprehensive National Science Center, Hefei 230088, China (e-mail:
xmxu@scut.edu.cn).

Digital Object Identifier 10.1109/TNNLS.2024.3400928}
}


\maketitle

\begin{abstract}
Low-rankness plays an important role in traditional machine learning, but is not so popular in deep learning. Most previous low-rank network compression methods compress networks by approximating pre-trained models and re-training. However, the optimal solution in the Euclidean space may be quite different from the one with low-rank constraint. A well-pre-trained model is not a good initialization for the model with low-rank constraints. Thus, the performance of a low-rank compressed network degrades significantly. Compared with other network compression methods such as pruning, low-rank methods attract less attention in recent years. In this paper, we devise a new training method, low-rank projection with energy transfer (LRPET), that trains low-rank compressed networks from scratch and achieves competitive performance. We propose to alternately perform stochastic gradient descent training and projection of each weight matrix onto the corresponding low-rank manifold. Compared to re-training on the compact model, this enables full utilization of model capacity since solution space is relaxed back to Euclidean space after projection. The matrix energy (the sum of squares of singular values) reduction caused by projection is compensated by energy transfer. We uniformly transfer the energy of the pruned singular values to the remaining ones. We theoretically show that energy transfer eases the trend of gradient vanishing caused by projection. In modern networks, a batch normalization (BN) layer can be merged into the previous convolution layer for inference, thereby influencing the optimal low-rank approximation of the previous layer. We propose BN rectification to cut off its effect on the optimal low-rank approximation, which further improves the performance. Comprehensive experiments on CIFAR-10 and ImageNet have justified that our method is superior to other low-rank compression methods and also outperforms recent state-of-the-art pruning methods. For object detection and semantic segmentation, our method still achieves good compression results. In addition, we combine LRPET with quantization and hashing methods and achieve even better compression than the original single method. We further apply it in Transformer-based models to demonstrate its transferability. Our code is available at https://github.com/BZQLin/LRPET.
\end{abstract}

\begin{IEEEkeywords}
Network Compression, Low-Rank Projection, Energy Transfer, Training Method.
\end{IEEEkeywords}

\section{Introduction}
\IEEEPARstart{C}{onvolutional} neural networks (CNNs) have attained state-of-the-art performance in computer vision applications, like image classification \cite{DBLP:conf/cvpr/HeZRS16,DBLP:journals/ijcv/RussakovskyDSKS15} and object detection \cite{DBLP:journals/pami/RenHG017}. 
However, such good performance is along with a large amount of storage and computing load, which restricts the deployment of CNN models on devices with limited resources. For instance, a ResNet-50 model needs roughly 4G FLOPs (floating-point operations per second) to categorize a color image with 224$\times$224 pixels.
Network compression is one kind of typical methods for handling redundant or insignificant network components. 
{Network compression includes
	low-rank decomposition \cite{DBLP:conf/iccv/WenXWWCL17,yu2017compressing,zhang2016accelerating,DBLP:conf/cvpr/KimKK19},
	network pruning (sparsity) \cite{DBLP:conf/cvpr/TungM18,DBLP:conf/ijcai/LinJZZW020,DBLP:conf/accv/LiCLSZ20,liu2018rethinking,liu2021discrimination}, parameter quantization \cite{DBLP:conf/cvpr/WuLWHC16,esser2019learned,lee2021network,zhuang2021effective}, knowledge distillation \cite{DBLP:journals/corr/HintonVD15}, network structure search\cite{guo2021towards}, hashing\cite{eban2020structured,obukhov2020t}, etc. Neural Architecture Transformer (NAT)\cite{guo2021towards} method proposes a network structure search method that casts the optimization problem into a Markov decision process. Learned Step Size Quantization (LSQ)\cite{esser2019learned} presents a parameter quantization method which introduce a novel means to estimate and scale the task loss gradient at each weight and activation layer’s quantizer step size. Element-wise gradient scaling (EWGS)\cite{lee2021network} further represents a simple yet effective alternative to the sub-optimal straight-through estimator. Effective Training of convolutional neural networks with low-bitwidth weights and activations \cite{zhuang2021effective} can further improve the training efficiency.  Structured multi-hashing (SMH) \cite{eban2020structured} combines ideas from weight hashing and
	dimensionality reductions resulting in a simple and powerful structured multi-hashing method based on matrix products. Obukhov et al.\cite{obukhov2020t} introduces a novel concept for a compact representation of a set of tensors call T-Basis, which can parameterize the tensor set with a small number of parameters. It is excited that these methods are not mutually exclusive, but can be combined with each other. Antonio et al.\cite{polino2018model} proposes a model compression method combining quantization and knowledge distillation.}

Among these methods, low-rankness and sparsity are important properties for analyzing the intrinsic structure of data in traditional machine learning \cite{DBLP:journals/tnn/GuoLXXT18,liu2013robust,he2011robust,wright2009robust,elhamifar2013sparse,yang2011robust}. However, when things come to network compression, low-rankness is overlooked by researchers compared to the popularity of pruning. Low-rankness and sparsity respectively capture the global and local information of data. Both of them should be important for network compression. {One popular method to train low-rank models is alternating gradient descent (AGD) \cite{agd}, which updates two blocks of variables in an alternating manner using gradient descent steps. When a low-rank matrix is decomposed into the product of two small matrices, AGD can be applied on the small matrices and thus solve low-rank problems. However, AGD only deals with two blocks of variables and cannot be directly applied on low-rank network decomposition for multiple layers.} In this paper, we investigate the obstacles to applying low-rankness to network compression and propose a training method to tackle these problems.

For low-rank approximation (LRA) methods \cite{DBLP:conf/bmvc/JaderbergVZ14}, reconstruction error may be accumulated layer by layer, degrading the performance even after fine-tuning. Since low-rank decomposed network and the original network are not of the same solution space, mimicking the original network is not the optimal choice. This can be verified by computing the distance between the merged weight of fine-tuned low-rank decomposed network and the weight of well-trained original architecture. We empirically find that the related distance ratio is usually more than 40\% for popular network architectures.
Thus, training low-rank compact network from scratch seems more appealing. However, it is a great challenge to achieve satisfactory results.
Low-rank network compression factorizes a large weight matrix into the product of small matrices, and thus the network becomes deeper by changing a layer into several consecutive small layers. When the model capacity is sufficient for the task, training a deeper network is usually harder. Trained rank pruning (TRP) \cite{DBLP:conf/ijcai/XuL0WWQCLX20} avoids training a deeper network
by performing low-rank projection (LRP) and training alternatively, which preserves and optimizes all parameters of the original network in the low-rank form. However, TRP needs computing singular value decomposition (SVD) at every iteration to guarantee its performance and is very time-consuming.
SVD training \cite{DBLP:conf/cvpr/YangTW0HLLC20} forces the factorized matrices in an SVD  form with orthogonal and sparsity regularization to ease the training. However, a low-rank network's capacity is limited compared with the original full-rank network, which makes it challenging to achieve comparable performance.

Batch normalization (BN), is often presented in conjunction with the convolutional layer, is a standard module in modern networks. After training, BN can be absorbed into the previous convolution layer for inference, which is equivalent to modifying its weight of the previous convolution layer. This will change the singular values of the weight matrix and the result of optimal low-rank projection. Thus, BN will affect the optimal LRA of the weight matrix for inference. However, past low-rank network compression methods has never considered the influence of the BN layers but directly projected the convolutional layer. We find that this negatively affects the projection, so we integrate the linear transformation of BN into the network weights in training and achieve significant performance improvements.

\begin{figure*}[t]
    \centering
    \includegraphics[width=0.65\textwidth]{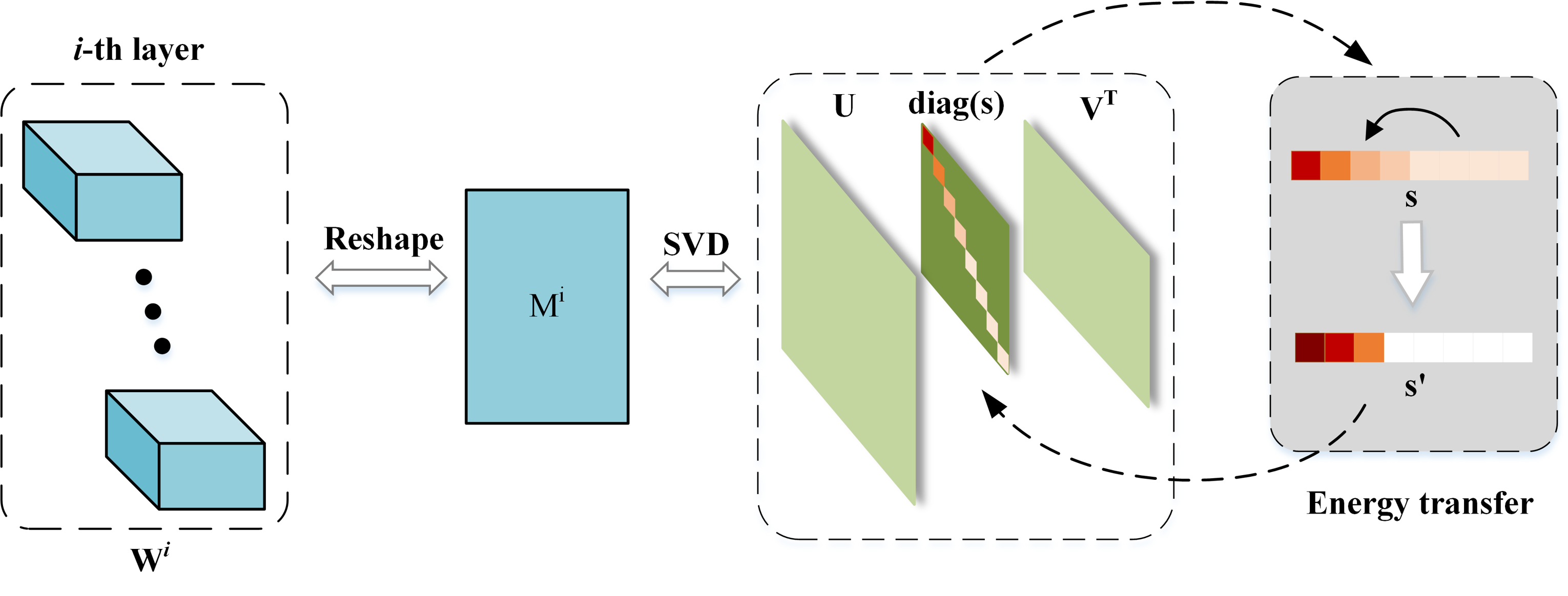}
    \caption{An illustration of the procedure of energy transfer. For each convolution layer, we reshape its weight data as matrix M\textsuperscript{i} and decompose it by SVD method, then we clip the singular value matrix diag(\textbf{s})  with energy transfer and finally replace the convolution layer weight data. The dotted arrow denotes extracting/recovering the singular values from/to the diagonal matrix. The solid arrow denotes transferring the energy.}
    \label{fig:energytrans} 
\end{figure*}

In this paper, we devise an efficient training method, low-rank projection with energy transfer (LRPET), that trains low-rank compressed networks from scratch and achieves competitive performance. Fig. \ref{fig:energytrans} illustrates the procedure of LRPET. 
We propose to alternately perform stochastic gradient descent training and projection of each weight matrix onto the corresponding low-rank manifold with matrix energy  (the sum of squares of singular values) transfer to compensate energy reduction caused by projection. Our contributions are summarized as follows:
%
\begin{itemize}

    \item LRP during training can lead to gradient vanishing. Through theoretical analysis, we find that this is a new problem caused by training method rather than the traditional gradient vanishing caused by network structure. We discover and solve this problem with LRPET. 
    
     \item Energy transfer is critical for training efficiency. To achieve good performance, TRP requires LRP  for every iteration. With energy transfer, LRPET only needs to apply LRP once for every epoch, significantly enhancing efficiency. 
    
    \item To the best of our knowledge, LRPET is the first method to deal with BN's effect in low-rank network compression. By merging the linear transformation of BN into network weights, LRPET is compatible with BN and improves the final performance.

\end{itemize}

Our method is exciting since we show that low-rankness has the potential to achieve comparable performance to sparsity in such a simple way, which is overlooked in the community of network compression.
We conduct a lot of experiments on the image classification task. The results show that our proposed method is effective and suitable for different networks (ResNet \cite{DBLP:conf/cvpr/HeZRS16},  GoogLeNet \cite{DBLP:conf/cvpr/SzegedyLJSRAEVR15}, VGG-16 \cite{DBLP:journals/corr/SimonyanZ14a}, FBNet\cite{wu2019fbnet} and GhostNet\cite{DBLP:conf/cvpr/HanW0GXX20})  ) on benchmark datasets (CIFAR-10 \cite{krizhevsky2009learning} and ImageNet \cite{DBLP:journals/ijcv/RussakovskyDSKS15}). To further analyze the generalization of our
method, we also apply it to object detection\cite{fasterrcnn} and semantic segmentation\cite{pspnet}. In addition, We combine our methods with LSQ \cite{esser2019learned} and SMH \cite{eban2020structured} , respectively, proving that our model has good scalability.

This paper is organized as follows. The related work is briefly reviewed in Section \ref{sec:relat}. Section \ref{sec:method} describes the details of the proposed LRPET method. Section \ref{sec:exper} presents experimental results, and Section \ref{sec:conclu} concludes the paper.

\section{Related works}
\label{sec:relat}

\subsection{Low-Rank Decomposition}
Low-rank decomposition methods decompose the weight matrix/tensor of a layer into several small matrices/tensors to reduce storage and computation cost. They can be divided into LRA and training low-rank networks from scratch.

\subsubsection{Low-Rank Approximation}
LRA methods achieve network compression by approximating pre-trained models and re-training.
Lebedev et al. \cite{DBLP:journals/corr/LebedevGROL14} decompose the 4-D convolution kernel tensor into the sum of rank-one tensors and replace the original convolutional layer with a sequence of four convolutional layers with small kernels. Nonetheless, increasing the number of layers can lead to gradient explosion issues, necessitating a carefully designed re-training process. Subsequent works \cite{DBLP:conf/bmvc/JaderbergVZ14, zhang2016accelerating} reshape the 4-D tensor into a 2-D matrix and  decompose the matrix into two small matrices, which results in two consecutive layers.
Jaderberg et al.\cite{DBLP:conf/bmvc/JaderbergVZ14} construct a low rank basis of rank-1
filters in the spatial domain by minimizing weight matrix reconstruction error and get two consecutive layers with convolutional kernel sizes of $1\times h$ and $w\times 1$. 
Zhang et al. \cite{zhang2016accelerating} reconstruct linear or non-linear response within an approximate low-rank subspace, which actually
divides a layer into two consecutive layers with convolution kernel sizes $w\times h $ and $1\times 1$.
However, the standalone performance of LRA may prove to be inadequate in practical applications. Consequently, recent works \cite{DBLP:journals/pami/LinJCTL19, DBLP:conf/cvpr/LiG0GT20} improve LRA by combining with other methods. 
Lin et al.\cite{DBLP:journals/pami/LinJCTL19} retrain the low-rank compression network through knowledge transfer and distillation. 

Collaborative compression (CC) \cite{DBLP:conf/cvpr/LiG0GT20} simultaneously learns the model sparsity and low-rankness to joints channel pruning and tensor decomposition.
However, combining methods usually introduces extra computation costs.

For the LRA methods, decomposing the pre-trained network may result in a significant loss of accuracy, and the network needs retraining to recover the performance as much as possible. A well-pre-trained model is not a good initialization for the model with low-rank constraints.
So some works try to obtain low-rank networks directly by training from scratch.

\subsubsection{Training From Scratch}
Training from scratch methods usually uses regularization to guide the training process to learn low-rank network structure.
Wen et al.\cite{DBLP:conf/iccv/WenXWWCL17} train neural networks toward lower-rank space by enforcing filters to become closer with a forced regularization motivated by physics.
TRP \cite{DBLP:conf/ijcai/XuL0WWQCLX20} alternates performing LRP and training, which maintains the capacity of the original network while imposing low-rank constraints. However, they apply LRP according to a pre-defined energy pruning ratio, which results in varying ranks across layers and destabilizes the training process. Although TRP argues that their method does not cause the problem of gradient vanishing, we theoretically find that gradient vanishing still exists. In this paper, we propose energy transfer to ease this problem and improve performance.
To guarantee the performance, TRP adds nuclear norm regularization, whose optimization needs time-consuming SVD at each iteration.
To avoid SVD's inefficiency during training, SVD training \cite{DBLP:conf/cvpr/YangTW0HLLC20} first decomposes each layer into the form of its full-rank SVD, and then performs training on the decomposed weight matrices. 
They use orthogonality and sparsity regularization to keep the singular vector matrices close to unitary matrices and sparsify the singular values.
To promote the performance of low-rank networks, Alvarez et al. \cite{DBLP:conf/nips/AlvarezS17} combine sparse group Lasso regularizer and nuclear norm-based low-rank regularizer, which encourages each layer's weight matrix to be filter-wise sparse and low-rank. 
{Efficient decomposition and pruning (EDP) \cite{DBLP:journals/tnn/RuanLYLHLM21} scheme minimizes the rank of the weight matrix and identifies the redundant channels by constructing a compressed-aware block.}
However, multiple regularizers require multiple trade-off hyperparameters, which are hard to tune.  Adding a regularizer means implicit constraint, and thus it is difficult to determine the compression rate before training. Note that a BN layer can be absorbed into the previous convolutional layer by modifying its weights after training and this will affect the optimal LRA of the weight matrix but has not been studied in previous low-rank network compression methods. In contrast, our LRPET is compatible with BN by applying LRP to the modified weight matrix that absorbs BN's parameter, and this effectively improves the performance.

\subsection{Network Pruning}

Network pruning, which is orthogonal to low-rank decomposition, aims to remove the redundant or unimportant parts of the neural network.

Non-structured pruning \cite{DBLP:conf/cvpr/TungM18} involves selectively removing individual weight elements within the network, often necessitating complex hardware and/or software implementations. Conversely, structured pruning \cite{DBLP:journals/corr/HanMD15} directly eliminates entire filters or blocks, making it simpler to implement and thus more prevalent in recent research\cite{DBLP:conf/ijcai/HeKDFY18, DBLP:conf/cvpr/Yu00LMHGLD18}.

Yang et al.\cite{yang2017designing} propose an energy-aware pruning algorithm that directly utilizes the energy consumption of a CNN to guide the pruning process. Li et al. \cite{DBLP:conf/iclr/0022KDSG17} removes whole filters with their connecting feature maps according to ${\ell}_1$-norm. Channel pruning \cite{DBLP:conf/iccv/HeZS17}  and ThiNet \cite{DBLP:journals/pami/LuoZZXWL19} prune the whole filters by minimizing layer output reconstruction error with different selection strategy: channel pruning uses LASSO regression while ThinNet adopts greedy algorithm. 
Different from the above methods that prune filters by only considering the statistics of an individual layer or two consecutive layers, Yu et al. \cite{DBLP:conf/cvpr/Yu00LMHGLD18} minimize the reconstruction error of important responses in the final response layer and propagate the importance scores of final responses to every filter in the network for pruning.
Considering that a smaller scaling parameter of  BN layer indicates that the corresponding convolution filter’s effect is more negligible, Liu et al. \cite{DBLP:conf/iccv/LiuLSHYZ17} imposes ${\ell}_1$ regularization on the scaling factors in BN, and channels with small scaling factors in BN layers are pruned. 
To prevent training inefficiency, HRank \cite{DBLP:conf/cvpr/LinJWZZ0020} investigates the rank of feature maps and proposes a new pruning criterion that prunes filters with low-rank feature maps. Aiming for even greater efficiency, Chen at al.\cite{chen2023rgp} propose a regular graph pruning (RGP) method to perform a one-shot neural network pruning.
However, these methods require fine-tuning to restore performance.

Soft Filter Pruning (SFP) \cite{DBLP:conf/ijcai/HeKDFY18} train pruned networks from scratch by alternatively pruning with $\ell_2$ norm and reconstructing the pruned filters from training. This soft manner can maintain the model capacity and avoid fine-tuning. Edropout\cite{edropout}  designs an energy-based model to stochastically evolve the population to find the best pruning state with lower energy loss. To further reduce redundancy, we propose an Ising energy model within an optimization framework for pruning convolutional kernels and hidden units. Learning filter pruning criteria (LFPC) \cite{DBLP:conf/cvpr/HeDLZZ020} develops a differentiable sampler to select pruning different criteria in pre-determined candidates for different layers.
Instead of pre-determining the pruning criterion, Lin et al. \cite{DBLP:conf/cvpr/LinJYZCYHD19} introduce a soft mask with sparsity regularization and generative adversarial learning (GAL) to select the pruned parts.
Deep compression with reinforcement learning (DECORE) \cite{DBLP:conf/cvpr/AlwaniWM22} automates the network compression process 
by assigning an agent to each channel along with a light policy gradient method to learn which neurons or channels to be kept or removed.
Fire together wire together (FTWT) \cite{DBLP:conf/cvpr/ElkerdawyE0R22} dynamically prune the network with a self-supervised binary classification module to predict a mask to process k filters in a layer based on the activation of its previous layer, which results in sample-wise adaptive pruned network structure. 
{Evolutionary algorithm-based method for shallowing deep neural networks at block levels (ESNB) \cite{DBLP:journals/tnn/ZhouYY22} automatically discover shallower network architectures by pruning less informative blocks through evolutionary search and employ knowledge distillation to recover the performance.}
FilterSketch \cite{DBLP:journals/tnn/LinCLYTLTJ22} encodes the second-order information of pre-trained weights and enables the representation capacity
of pruned networks to be recovered with a simple fine-tuning
procedure.
EPruner \cite{DBLP:journals/tnn/LinJLWWHY22} introduce a graph message-passing algorithm Affinity Propagation \cite{frey2007clustering} to CNNs to figure out an adaptive number of the most important filters. Rather than utilizing some indirect criteria to guide connection or channel pruning, Chen et al.\cite{dcph} design criteria directly related to the final accuracy of a network called conditional accuracy changes to evaluate the importance of each channel. Instead of using the same pruning rate across all layers, Particle Swarm Optimization(PSO)\cite{pso} proposes a network redundancy elimination approach that employs particle swarm optimization to learn pruning rates in a layerwise manner. Ganesh et al.\cite{snacs} combine a probabilistic pruning framework with constraints on the underlying weight matrices and propose a fast hash-based estimator of adaptive conditional mutual information. 
These methods benefit from the recently developed learning techniques, but at the same time increase the complexity of the training process.

\section{Low-Rank Projection with Energy Transfer}
\label{sec:method}
A deep neural network composed of $L$ layers is to successively compute the features
$\mathbf{x}^l=\phi(\mathbf{W}^l \mathbf{x}^{l-1}+\mathbf{b}^{l})\in\mathbb{R}^{N_{l}}\,\forall l=1,2,\cdots, L$,
where $\mathbf{x}^{l-1}\in \mathbb{R}^{N_{l-1}}$ is the input feature of the $l^{th}$
layer, $\phi(\cdot)$ is element-wise nonlinear activation function, 
and $\mathbf{W}^{l}\in \mathbb{R}^{N_{l}\times N_{l-1}}$ 
and $\mathbf{b}^{l}\in \mathbb{R}^{N_{l}}$ are learnable weight matrix and bias vector, respectively. Note that through reshaping weight and input tensors into matrices in a proper way, convolution can be implemented by matrix multiplication.
Thus, we use a weight matrix to denote either full connection or convolution. Although the input and output need reshape operation, we ignore them in the formulation for simplification since this does not affect the analysis in this paper. Suppose the final composition function of all layers is $f(\mathbf{x};\Theta)$, where $\Theta=\{\mathbf{W}^{l},\mathbf{b}^{l}|l=1,2,\cdots,L\}$ denotes the set of learnable parameters. Given the training dataset $D=\{(\mathbf{x}_{i},\mathbf{y}_{i})\}_{i=1}^{M}$ composed of $M$ samples with inputs $\mathbf{x}_{i}\in\mathbb{R}^{N_{0}}$ and outputs $\mathbf{y}_{i}\in\mathbb{R}^{N_{y}}$, and the loss function $\mathcal{L}(\mathbf{y},f(\mathbf{x};\Theta))$, the optimization objective function of an ordinary deep neural network is given by $\min_{\Theta} \mathbb{E}_{(\mathbf{x},\mathbf{y})\in D}[\mathcal{L}(\mathbf{y},f(\mathbf{x};\Theta))]$.

\subsection{Overview}
In this section, we introduce the proposed method LRPET and its relevant operations. Since LRPET is a variant of low-rank projection by adding energy transfer and BN rectification, we introduce these components sequentially. Then, we discuss how to set the rank of each layer when applying LRPET. Finally, we show how to inference from the network after training.
\subsection{Low-Rank Projection}
A low-rank matrix can be factorized into the product of two small matrices to save parameters and computation cost in matrix multiplication. Thus, learning low-rank weight matrices for each layer can compress the networks. 
Previous low-rank compression methods \cite{yu2017compressing,zhang2016accelerating} usually need pre-training, decomposition, and re-training to achieve this goal, which is time-consuming. In practice, the weights of neural networks are often nearly full-rank rather than exactly low-rank. This observation is supported by a statistical experiment presented in the AAAI 2023 paper [1]. According to the Eckart-Young-Mirsky theorem \cite{markovsky2012low}, for a matrix, its rank-k projection via Singular Value Decomposition (SVD) has the closest Euclidean distance among all rank-k matrices. The optimal solution with low-rank constraints may be quite different from the one in the Euclidean space, and thus the pre-trained model is hard to guarantee good initialization for low-rank matrices. 
In this paper, we propose a training from scratch method to learn low-rank compressed models. The problem is formulated as follows:
\begin{equation}
    \begin{array}{ll}
        \min_{\Theta} \mathbb{E}_{(\mathbf{x},\mathbf{y})\in D}[\mathcal{L}(\mathbf{y},f(\mathbf{x};\Theta))]\\
        s.t.\quad \mathbf{W}^{l}\in\Omega_{r_{l}}^{N_{l}, N_{l-1}}\, \forall l \in\{1,2,\cdots,L\},
    \end{array}
\end{equation}
where $\Omega_{r_{l}}^{N_{l}, N_{l-1}}$ stands for the manifold of rank-$r_{l}$ matrices with the size of $N_{l}\times N_{l-1}$. 

Intuitively, this problem can be solved by projected stochastic gradient descent (SGD), i.e., one step SGD update followed by projection onto the corresponding low-rank manifold. Since the operation for each layer is the same, we respectively simplify $\mathbf{W}^{l}$ and $\Omega_{r_{l}}^{N_{l}, N_{l-1}}$ to $\mathbf{W}$ and $\Omega_{r}$ in the following.
According to Eckart-Young theorem \cite{xiang2012optimal}, the projection $\mathcal{P}_{\Omega_{r}}(\mathbf{W})=\mathop{\arg\min}\limits_{\hat {\mathbf{W}}\in \Omega_{r}}\|\hat{\mathbf{W}}-\mathbf{W}\|_{F}^{2}$ can be computed by 
\begin{equation}\label{LR_proj}
    \mathcal{P}_{\Omega_{r}}(\mathbf{W})=\sum_{i=1}^{r}s_{i}\mathbf{u}_{i}\mathbf{v}_{i}^{T},
\end{equation}
where $s_{i}$ is the $i^{th}$ largest singular value of $\mathbf{W}$, $\mathbf{u}_{i}$ and $\mathbf{v}_{i}$ are the corresponding singular vectors of $\mathbf{U}$ and $\mathbf{V}$ in the SVD $\mathbf{W}=\mathbf{U}\mathbf{S} \mathbf{V}^{T}$. However, applying SVD for $\mathbf{W}$ after every iteration is time-consuming. Motivated by singular value bounding (SVB) \cite{jia2017improving} that bounds the singular value after a certain number of iterations, we use a variant of projected SGD that projects $\mathbf{W}$ onto $\Omega_{r}$ after a certain number of iterations, which does not increase too much training time. 

\subsection{Energy Transfer}\label{ET}
LRP is a non-expansive operator that will reduce not only the Frobenius norm of the weight matrix but also the norm of the gradient of the corresponding feature map when back propagation. The subscripts ``$1$$:$$r$$,:$'' and ``$:,$$1$$:$$r$'' denote the first $r$ rows and the first $r$ columns of the matrix, respectively. $\mathbf{U}_{\perp}$ denotes the orthogonal complement of $\mathbf{U}_{:,1:r}^{T}$, i.e., the remainder columns of $\mathbf{U}$. LRP can also be computed by 
\begin{equation}
    \mathcal{P}_{\Omega_{r}}(\mathbf{W})=\mathbf{U}_{:,1:r}\mathbf{U}_{:,1:r}^{T}\mathbf{W}=\mathbf{W}\mathbf{V}_{:,1:r}\mathbf{V}_{:,1:r}^{T}.
\end{equation}
Then, we have
\begin{equation}
    \begin{array}{lll}
        \|\mathcal{P}_{\Omega_{r}}(\mathbf{W})\|_{F}^{2}&=&\|\mathbf{U}_{:,1:r}\mathbf{U}_{:,1:r}^{T}[\mathbf{U}_{:,1:r},\mathbf{U}_{\perp}]\mathbf{S} \mathbf{V}^{T}\|_{F}^{2}\\
        &=&\|[\mathbf{U}_{:,1:r},\mathbf{0}]\mathbf{S} \mathbf{V}^{T}\|_{F}^{2}\\
        &=&\|\mathbf{U}_{:,1:r}(\mathbf{S} \mathbf{V}^{T})_{1:r,:}\|_{F}^{2}\\
        &=&\|(\mathbf{S} \mathbf{V}^{T})_{1:r,:}\|_{F}^{2},
    \end{array}
\end{equation}
where $\mathbf{0}$ is a zero matrix. 
Note that $\|W\|_{F}^{2}=\|SV^{T}\|_{F}^{2}$ and the square of Frobenius norm of a sub-matrix is usually smaller than the one of a whole matrix. Thus, the energy is likely to reduce after projection. 

Let $\mathbf{x}$ denote the input data of a layer and $\hat{\mathbf{h}}=\mathbf{W}\mathbf{x}$. Then, the gradient $\frac{\partial \mathcal{L}}{\partial \mathbf{x}}$ is given by $\frac{\partial \mathcal{L}}{\partial \mathbf{x}}=W^{T}\frac{\partial \mathcal{L}}{\partial \hat{\mathbf{h}}}$. After projection, the energy of the gradient becomes
\begin{equation}
    \|\frac{\partial \mathcal{L}}{\partial \mathbf{x}}\|^{2}=\|\mathcal{P}_{\Omega_{r}}(\mathbf{W})^{T}\frac{\partial \mathcal{L}}{\partial \hat{\mathbf{h}}}\|_{F}^{2}=\|\mathbf{V}_{:,1:r}\mathbf{V}_{:,1:r}^{T}\mathbf{W}^{T}\frac{\partial \mathcal{L}}{\partial \hat{\mathbf{h}}}\|_{F}^{2}.
\end{equation}

By similar deduction, we can conclude that the gradient energy $\|\frac{\partial \mathcal{L}}{\partial \mathbf{x}}\|$ is usually reduced after projection. 
By back propagation, the gradient of the trainable parameters of previous layer also decreases.
Thus, we tend to face the problem of gradient vanishing as the network goes deeper. 

To prevent gradient vanishing, it is necessary to augment the gradient energy after projection. Note that SVD splits a system into a set of linearly independent components and the singular values represent the square root of the energy of the corresponding component \cite{sadek2012svd}. The gradient energy reduction is due to the loss of weight matrix energy after projection. To get rid of gradient energy reduction, we propose to maintain the energy of the weight matrix with a simple solution that transfers the energy of the less important components to the more important ones instead of cutting them off. We use $\mathbf{s}$ to denote the vector composed of the singular values of $W$ in descending order, i.e., $\mathbf{s}=\text{diag}(\mathbf{S})$, and use $\mathbf{s}_{1:r}$ to denote the first $r$ elements of $\mathbf{s}$. To compensate for the reduced energy, we can update the first $r$ singular values by multiplying them with a coefficient
\begin{equation}
    \alpha = \|\mathbf{s}\|/\|\mathbf{s}_{1:r}\|.
\end{equation}
Then, we replace the singular values in Eq. (\ref{LR_proj}) with the updated ones, which can be equivalently represented by  $\alpha\mathcal{P}_{\Omega_{r}}(\mathbf{W})$. 

If the training dataset is sufficiently large and diverse, and the weights are initialized with zero mean and appropriate variance, the gradient of feature maps tends to balance out in different directions, leading to an expectation of zero. In modern CNNs, BN is a popular module that ensures the activations have similar variances and tends to reduce the covariance between different dimensions. This will leads to more stable gradients, i.e., the gradients are less likely to exhibit extreme variations or strong correlations. Thus, we can approximately assume that the covariance matrix of the gradient of feature maps is diagonal with equal variances along all dimensions. Note that these assumptions are also make in the work orthogonalization method using Newton’s iteration (ONI) \cite{DBLP:conf/cvpr/Huang00WYL020} to study the effect of orthogonality on gradient of feature maps. The performance of ONI is consistent with the theoretical analysis based on these assumptions, which also suggests that the reasonableness of these assumptions.
Based on these assumptions, we have the following theorem for the expectation of gradient energy and gradient norm.
\newtheorem{theorem}{Theorem}
\begin{theorem}
\label{theorem1}
    $\mathbf{W}$ and $\mathbf{x}$ represent the learnable weight matrix and input data of a layer in the network, respectively.
    Let $\hat{\mathbf{h}}=\mathbf{W}\mathbf{x}$.
    Assume that $\mathbb{E}(\frac{\partial \mathcal{L}}{\partial \hat{\mathbf{h}}})=\mathbf{0}$ and
    $\text{cov}(\frac{\partial \mathcal{L}}{\partial \hat{\mathbf{h}}})=\sigma^{2}\mathbf{I}$, 
    	where $\mathbf{0}$ is a zero matrix, $\mathbf{I}$ is an identity matrix, and $\sigma>0$ is a constant. 
    We have $\mathbb{E}(\|\frac{\partial \mathcal{L}}{\partial \mathbf{x}}\|^2)=\sigma^{2}\|\mathbf{W}\|_{F}^2$ and $\mathbb{E}(\|\frac{\partial \mathcal{L}}{\partial \mathbf{x}}\|)\leq \sigma\|\mathbf{W}\|_{F}$.
\end{theorem}
\begin{IEEEproof}
	Given $\hat{\mathbf{h}}=\mathbf{W}\mathbf{x}$, the gradient $\frac{\partial \mathcal{L}}{\partial \mathbf{x}}$ can be computed by
	\begin{equation}
	\frac{\partial \mathcal{L}}{\partial \mathbf{x}}=\mathbf{W}^{T}\frac{\partial \mathcal{L}}{\partial \hat{\mathbf{h}}}.
	\end{equation} 
	Since $\mathbb{E}(\frac{\partial \mathcal{L}}{\partial \hat{\mathbf{h}}})=0$ and $\text{cov}(\frac{\partial \mathcal{L}}{\partial \hat{\mathbf{h}}})=\sigma^2\mathbf{I}$, we have
	\begin{equation}
	\begin{aligned}
	\mathbb{E} [ (\frac{\partial \mathcal{L}}{\partial \hat{\mathbf{h}}})(\frac{\partial \mathcal{L}}{\partial \hat{\mathbf{h}}})^T ] &=
	\mathbb{E} [ (\frac{\partial \mathcal{L}}{\partial \hat{\mathbf{h}}} - \mathbb{E}(\frac{\partial \mathcal{L}}{\partial \hat{\mathbf{h}}}))(\frac{\partial \mathcal{L}}{\partial \hat{\mathbf{h}}} - \mathbb{E}(\frac{\partial \mathcal{L}}{\partial \hat{\mathbf{h}}}))^T ] \\&=\text{cov}(\frac{\partial \mathcal{L}}{\partial \hat{\mathbf{h}}})=\sigma^2\mathbf{I}.
	\end{aligned}
	\end{equation}
	The expectation of the square of the gradient norm becomes
	\begin{equation}
	\begin{array}{lll}
	\mathbb{E}(\|\frac{\partial \mathcal{L}}{\partial \mathbf{x}}\|^2) 
	&= \mathbb{E}(tr( \frac{\partial \mathcal{L}}{\partial \mathbf{x}} \frac{\partial \mathcal{L}^T}{\partial \mathbf{x}}) )
	\\&= \mathbb{E}(tr( \mathbf{W}^{T}\frac{\partial \mathcal{L}}{\partial \hat{\mathbf{h}}} \frac{\partial \mathcal{L}^T}{\partial \hat{\mathbf{h}}} \mathbf{W}))
	\\&= \mathbb{E}(tr( \frac{\partial \mathcal{L}}{\partial \hat{\mathbf{h}}} \frac{\partial \mathcal{L}^T}{\partial \hat{\mathbf{h}}} \mathbf{W}\mathbf{W}^{T}))
	\\&= tr( \mathbb{E}(\frac{\partial \mathcal{L}}{\partial \hat{\mathbf{h}}} \frac{\partial \mathcal{L}^T}{\partial \hat{\mathbf{h}}} \mathbf{W}\mathbf{W}^{T}))
	\\&= tr( \mathbb{E}(\frac{\partial \mathcal{L}}{\partial \hat{\mathbf{h}}} \frac{\partial \mathcal{L}^T}{\partial \hat{\mathbf{h}}}) \mathbf{W}\mathbf{W}^{T})
	\\&=tr(\sigma^2\mathbf{W}\mathbf{W}^{T})
	\\&=\sigma^2tr(\mathbf{W}\mathbf{W}^{T})
	\\&=\sigma^2\|\mathbf{W}\|^2_{F},
	\end{array}
	\end{equation}
	where $tr(\cdot)$ indicates the trace of the corresponding matrix. 
	We use $\mathbb{V}(\cdot)$ to denote the variance of the value. As it is well known, the variance can be represented as follows:
	\begin{equation}
	\begin{aligned}
	\mathbb{V}(\|\frac{\partial \mathcal{L}}{\partial \mathbf{x}}\|)=\mathbb{E}[(\|\frac{\partial \mathcal{L}}{\partial \mathbf{x}}\|)^2] -  \mathbb{E}[(\|\frac{\partial \mathcal{L}}{\partial \mathbf{x}}\|)]^2.
	\end{aligned}
	\end{equation}
	Because $\mathbb{V}(\|\frac{\partial \mathcal{L}}{\partial \mathbf{x}}\|) \geq 0$, we have the following relationship for the expectation of the gradient norm
	\begin{equation}
	\begin{aligned}
	\mathbb{E}(\|\frac{\partial \mathcal{L}}{\partial \mathbf{x}}\|) &\leq \sqrt{ \mathbb{E}[(\|\frac{\partial \mathcal{L}}{\partial \mathbf{x}}\|)^2]}
	\\&\leq \sqrt{ \sigma^2\|\mathbf{W}\|^2_{F} }
	\\&\leq  \sigma\|\mathbf{W}\|_{F} .
	\end{aligned}
	\end{equation}
\end{IEEEproof}

Compare to ONI \cite{DBLP:conf/cvpr/Huang00WYL020}, we have less assumptions and constraints on the weight matrix, and our conclusion is about the relationship between input and weight rather than input and output. 
Theorem \ref{theorem1} gives a corollary that low-rank projection will decrease the norm of the gradient of input feature maps. By back propagation, this will also decreases the norm of the gradient of the weight matrix in the previous layer. Note that low-rank projection is applied on multiple layers, after the accumulation of the gradient decreasing effect layer by layer, the gradient will become smaller.  
Theorem \ref{theorem1} also shows that the expectation of gradient energy and the upper bound of the expectation of gradient norm are kept if the Frobenius norm of $\mathbf{W}$ is the same. Note that $\|\alpha\mathcal{P}_{\Omega_{r}}(\mathbf{W})\|_{F}=\|\mathbf{W}\|_{F}$. Our energy transfer strategy can keep the gradient energy and eases the trend of gradient vanishing.

\subsection{BN Rectification}
BN, which inserts a learnable layer to normalize each layer's neuron activations as zero mean and unit variance, is popular in modern CNN. After training,  the combination of a BN layer and its previous convolutional layer is equivalent to a new convolutional layer. This changes the convolution weight and will affect the optimal low-rank approximation.
To analyze the effect of BN, we reformulate BN in the following:
\begin{equation}\label{bn_merge}
\begin{array}{lll}
\text{BN}(\mathbf{W}\mathbf{x})&=& \mathbf{\Gamma} \mathbf{\Sigma}^{-1} (\mathbf{W}\mathbf{x}-\mathbf{\bm{\mu}})+\mathbf{\bm{\beta}}\\
&=&\mathbf{\Gamma} \mathbf{\Sigma}^{-1}\mathbf{W}\mathbf{x}+(\mathbf{\bm{\beta}}-\mathbf{\Gamma} \mathbf{\Sigma}^{-1}{\bm{\mu}}),
\end{array}
\end{equation}
where $\mathbf{\Gamma} \in \mathbb{R}^{N \times N}$ and $\mathbf{\Sigma} \in \mathbb{R}^{N \times N}$ are diagonal matrices containing trainable scalar parameters and the neuron-wise output standard deviation (a small constant is added for numerical stability), respectively, $\bm{\mu} \in \mathbb{R}^{N}$ is the  mean output of the $N$ neurons in the layer, and $\bm{\beta} \in\mathbb{R}^{N}$ is a trainable bias term. Thus, after absorbing BN, the new convolution weight is $\mathbf{\widetilde{W}}=\mathbf{\Gamma} \mathbf{\Sigma}^{-1}\mathbf{W}$. Suppose the $i^{th}$ singular value of a matrix $\mathbf{A}$ is $\sigma_{i}(\mathbf{A})$. We have 
\begin{equation}
\sigma_{i}(\mathbf{\widetilde{W}})\leq \sigma_{i}(\mathbf{\Gamma} \mathbf{\Sigma}^{-1})\sigma_{i}(\mathbf{W})
\end{equation}
The singular values of $\mathbf{\widetilde{W}}$ are usually different from $\mathbf{W}$.
Thus, when taking BN into account, LRPET cannot be directly applied to $\mathbf{W}$.

We apply our energy transfer strategy to $\mathbf{\widetilde{W}}$ and study how to update the corresponding $W$. Denote the corresponding matrix of $\mathbf{\widetilde{W}}$ after energy transfer as $\mathbf{\widetilde{W}}'$.
To figure out the estimation $\hat {\mathbf{W}}$ of the original weight matrix $\mathbf{W}$, we formulate it as the following optimization problem:
\begin{equation}
\min\|\mathbf{\Gamma} \mathbf{\Sigma}^{-1}\hat{\mathbf{W}}-\mathbf{\widetilde{W}}'\|_F^2
\end{equation}
By letting the gradient be zero, we obtain the closed solution
\begin{equation}
    \hat{\mathbf{W}}=[ (\mathbf{\Gamma} \mathbf{\Sigma}^{-1})^T\mathbf{\Gamma} \mathbf{\Sigma}^{-1}]^{-1}(\mathbf{\Gamma} \mathbf{\Sigma}^{-1})^T\mathbf{\widetilde{W}}'.
\end{equation}
To avoid computing the inverse of singular matrix, we modify the solution to 
\begin{equation}
    \hat{\mathbf{W}}=[ (\mathbf{\Gamma} \mathbf{\Sigma}^{-1})^T\mathbf{\Gamma} \mathbf{\Sigma}^{-1}+ {\epsilon}]^{-1}(\mathbf{\Gamma} 
 \mathbf{\Sigma}^{-1})^T\mathbf{\widetilde{W}}',
\end{equation}
where $\epsilon>0$ is a small scalar and is set to 1e-5 in this paper. This process is called BN rectification (BNR).
The proposed method is summarized in Algorithm \ref{alg1}. 

\begin{algorithm}[t]
    \caption{Low-Rank Projection with Energy Transfer}
    \label{alg1}
    \begin{algorithmic}[1]
        \REQUIRE ~~\\ The parameters $\Theta=\{\mathbf{{W}}^{l},\mathbf{{b}}^{l}, \mathbf{\Gamma}^l,\bm{\beta}^{l}|l=1,2,\cdots,L\}$ of a network of $L$ convolutional and BN layers, the neuron-wise output standard deviation matrices $\mathbf{\Sigma} \in \mathbb{R}^{N \times N}$  of the $L$ BN layers, the desired ranks $\{r_{l}|l=1,2,\cdots,L\}$ of the $L$ layers, 
        the total number of training iterations $N$, and
        a specified number $T$ of iteration steps. 
        \FOR{$n=1,\cdots, N$}
        \STATE Update the parameters using SGD for one iteration.
        \WHILE{training proceeds for every $T$ interations}
        \FOR{$l=1, \cdots, L$}
        \STATE Compute $\mathbf{\widetilde{W}}^{l}={{\mathbf{\Gamma}}^l}{(\mathbf{\Sigma}^{-1})^{l}} \mathbf{W}^l$.
        \STATE Compute the SVD $\mathbf{\widetilde{W}}^{l}\overset{\text{SVD}}{=}\mathbf{U}^{l} \mathbf{S} ^{l} (\mathbf{V}^{l})^{T}$
        \STATE Compute $\alpha^{l}=\|\mathbf{s}^{l}\|/\|\mathbf{s}^{l}_{1:r_{l}}\|$ with $\mathbf{s}^{l}=\text{diag}(\mathbf{S}^{l})$.
        \STATE Compute $\hat{s}_{i}^{l}=\alpha^{l}s_{i}^{l}\,\forall 1,\cdots,r_{l}$.
        \STATE Update the first $r_{l}$ diagonal entries of $S^{l}$ with $\{\hat{s}_{i}^{l}\}_{1}^{r_{l}}$ and set the remainders to zeros.
        \STATE Update $\mathbf{{W}}^{l}$ with\\ 
         \begin{footnotesize}$[ ({{\mathbf{\Gamma}}^l}{(\mathbf{\Sigma}^{l})^{-1}})^T{{\mathbf{\Gamma}}^l}{(\mathbf{\Sigma}^{l})^{-1}}+ {\epsilon}]^{-1}({{\mathbf{\Gamma}}^l}{(\mathbf{\Sigma}^{l})^{-1}})^T \mathbf{U}^{l}\mathbf{S} ^{l}(\mathbf{V}^{l})^{T}$. \end{footnotesize}
        
        \ENDFOR
        \ENDWHILE
        \ENDFOR
    \end{algorithmic}
\end{algorithm}

\subsection{Rank Setting}
\label{sec:hyper}
Here we propose two rank setting methods: one is simple and effective method with only one hyperparameter, the other is advanced hyperparameter search method based on Bayesian optimization.

\subsubsection{Unique Rank Pruning Ratio}
The ranks $r_{l}$'s are hyperparameters and it is hard to set them layer by layer. We introduce a pruning ratio $p_{l}\in[0,1]$ to determine the rank for the $l^{th}$ layer. Mathematically, $r_{l}$ is given by $(1-p_{l})\min(N_{l},N_{l-1})$. Then, the problem changes to how to set $p_{l}$'s. We find that manually setting $p_{l}$'s to the same value for all layers works well in practice. Thus, we only need one hyperparameter to set the rank. In the following, we use $P$ to denote the unique rank pruning ratio.

\subsubsection{Bayesian Optimization}
To set the ranks of different layers distinguishably for better performance, we resort to Bayesian optimization. It is a general framework for minimizing blackbox objective functions that are
expensive to evaluate\cite{DBLP:conf/ijcai/WangZHMF13}.
The goal is to trade off between network performance and model size. The objective function is given by
\begin{equation}
    \min_{\mathbf{p}} \mathcal{E}(\mathbf{p})+\lambda C(\mathbf{p}),
\end{equation}
where $\mathbf{p}=[p_{1},p_{2},\cdots, p_{L}]$, $\mathcal{E}(\mathbf{p})$ is the top-1 error of the compressed model,  $C(\mathbf{p})$ is the compression rate, and $\lambda$ is the trade-off parameter. $C(\mathbf{p})$ is defined by $C(\mathbf{p})=D(\mathbf{p})/D_{ori}$, where $D(\mathbf{p})$ and $D_{ori}$ are respectively the sizes of the compressed and the original models. We model the objective function as a Gaussian process and use the expected improvement for evaluation to select the most promising candidate.  We refer readers to the work \cite{DBLP:conf/cvpr/TungM18} for more details.  
To save searching time, the top-1 error $\mathcal{E}(\mathbf{p})$ can be  estimated by training with much fewer epochs than ordinary training.

Note that the Bayesian optimization is not part of LRPET. We just show that our method is compatible to hyper parameters (rank ratios) search. We should emphasize that our method works well with a unique rank ratio without searching. Without confusion, we will use LRPET-S to denote LRPET with Bayesian optimization for searching.

\subsection{Network Inference}
After training by LRPET,
the singular values outside the range $[1,r_{l}]$ are near zero at each layer.	We can set these small singular values to zeros without performance degeneration. For inference, an ordinary convolutional layer changes into two cascaded small layers. 
Given the skinny SVD $\mathbf{W}=\mathbf{U} \mathbf{S} \mathbf{V}{^{T}}$, we construct the weight matrices of the two layers with $\mathbf{U}\sqrt{\mathbf{S}}$ and $\sqrt{\mathbf{S}}\mathbf{V}{^{T}}$. Suppose  $\mathbf{W}\in \mathbb{R}^{m\times n}$ and the rank is $r$. The number of parameters changes from $mn$ to $(m+n)r$, and the computational complexity of $\mathbf{W}\mathbf{x}$ also changes from $O(mn)$ to $O((m+n)r)$. If $r$ is small enough, such factorization will greatly reduce parameters and computational cost.

\section{Experiments}
\label{sec:exper}
In this section, LRPET is evaluated on two benchmark image classification datasets CIFAR-10 \cite{krizhevsky2009learning} and ImageNet \cite{DBLP:journals/ijcv/RussakovskyDSKS15}. Several mainstream CNN models are taken into account, including ResNets \cite{DBLP:conf/cvpr/HeZRS16}, GoogLeNet \cite{DBLP:conf/cvpr/SzegedyLJSRAEVR15}, and VGG-16 \cite{DBLP:journals/corr/SimonyanZ14a}.  Note that VGG-16 is a modified version of the original following \cite{DBLP:conf/cvpr/LinJWZZ0020}. 
Our experiments are implemented  with PyTorch \cite{DBLP:conf/nips/PaszkeGMLBCKLGA19}. 
\subsection{Datasets and Experimental Setting}
\subsubsection{Datasets}
{The CIFAR-10 dataset comprises 60,000 color images, each with a resolution of 32×32 pixels. These images are evenly distributed across 10 classes, with 5,000 images designated for training and 1,000 for testing per class.
We train the network according to standard data enhancement \cite{DBLP:journals/cacm/KrizhevskySH17,DBLP:conf/eccv/ZeilerF14}. For training, each image side is first zero-padded with 4 pixels, and then a 32$\times$32 patch is randomly cropped from the padded image or its horizontal flip with a probability of 0.5. We directly use the original 32$\times$32 images for test.
There are 1.28 million training images and 50,000 validation images of 1,000 classes in ImageNet dataset \cite{DBLP:journals/ijcv/RussakovskyDSKS15}.
Default ImageNet setting of PyTorch for data augmentation is adopted.}

\subsubsection{Experimental Setting}
On CIFAR-10, training is based on SGD with a batch size of 128, a momentum of 0.9, and a weight decay of 5e-4.
The initial learning rate is set to 0.1 and is divided by ten at the 50\% and 75\% of the total epoch number. 
We train the networks from scratch with LRPET for 400 epochs. The results of baselines of quoted from CC \cite{DBLP:conf/cvpr/LiLLYWC0M0J21}.
On ImageNet, networks are trained from scratch with LRPET for 120 epochs a batch size of 256, a momentum of 0.9, and a weight decay of 5e-4. The initial learning rate is set to  0.1 and is divided by 10 at epochs 30, 60, and 90. 

Note that a lot of compared methods use pre-trained model. For fair comparison, we also conduct experiments that first use the pre-trained models as initialization and then apply LRPET. On CIFAR-10, the pre-trained model is taken from CC \cite{DBLP:conf/cvpr/LiLLYWC0M0J21} or trained by ourselves with SGD for 164 epochs. The other training settings are the same as above. The pre-trained models on ImageNet are taken from PyTorch. When applying LRPET, the training epoch numbers for CIFAR-10 and ImageNet are 200 and 120, respectively. {The other training settings are the same as training from scratch except that the initial learning rate is set to 0.01.}  
In the following experimental results, we will use ``Y'' and ``N'' in a ``PT?'' column to indicate whether using  pre-trained model or not, respectively.

We set the iteration interval $T$ to one epoch on CIFAR-10 and  500 iterations for ImageNet. All the experiments of LRPET on CIFAR-10 are repeated for three times.
For rank ratio searching, the iteration number of Bayesian optimization is set to 20 and the trade-off parameter $\lambda$ is set to 1. During the evaluation of classification error for Bayesian optimization, each model undergoes 60 epochs of training, with the learning rate being divided by 10 every 20 epochs. Specifically, since searching for ImageNet is too time consuming, we search the hyperparameters on CIFAR-10 and then apply it to ImageNet. 

\subsection{Image Classification Results}
\subsubsection{Results on CIFAR-10}
We compare LRPET with other low-rank compression and network pruning methods by conducting experiments on CIFAR-10 with ResNets, GoogLeNet, and VGG-16.

\begin{table}[t]
	\footnotesize
	\centering
	\tabcolsep=0.5pt
	\caption{Results of ResNet-18 on CIFAR-10}
	\label{cifar_resnet18}
	\scalebox{1}{
		\centering
		\begin{tabular}{l|c|c|c|c|c}
			\hline
			Method         &PT?   & \tabincell{c}{Acc.(\%)} & $\uparrow \downarrow$(\%) & FLOPs (PR) & Params (PR)  \\
			
			\hline
			Baseline &N & 92.81 &0.00$\downarrow$ &556.65M(0.0\%) &11.17M(0.0\%) \\
			\hdashline
			EDropout \cite{edropout} &N  & 90.96  &1.85$\downarrow$ & - & 5.55M (50.3\%) \\
                IPruning \cite{ising} &N  & 84.09  &8.72$\downarrow$ & - & 5.49M (50.8\%) \\
			LRPET ($P$=0.58) &N & 92.05$\pm$0.06 &\textbf{0.76$\downarrow$} &249.42M (55.2\%)&\textbf{5.05M (54.8\%)}\\	
			
			\hline
	\end{tabular}}
\end{table}

\begin{table}[t]
	\footnotesize
	\centering
	\tabcolsep=0.5pt
	\caption{Results of ResNet-50 on CIFAR-10}
	\label{cifar_resnet50}
	\scalebox{1}{
		\centering
		\begin{tabular}{l|c|c|c|c|c}
			\hline
			Method         &PT?   & \tabincell{c}{Acc.(\%)} & $\uparrow \downarrow$(\%) & FLOPs (PR) & Params (PR)  \\
			
			\hline
			Baseline &N & 92.21 &0.00$\downarrow$ &1304.70M(0.0\%) &23.52M(0.0\%) \\
			\hdashline
			EDropout \cite{edropout} &N  & 85.30  &6.91$\downarrow$ & - & 10.91M (54.6\%) \\
                IPruning \cite{ising} &N  & 82.45  &9.76$\downarrow$ & - & 10.22M (56.5\%) \\
			LRPET ($P$=0.70) &N & 91.66$\pm$0.15 &\textbf{0.55$\downarrow$} &480.45M (63.2\%)&\textbf{8.60M (63.4\%)}\\	
			
			\hline
	\end{tabular}}
\end{table}

\begin{table}[t]
	\footnotesize
    \centering
    \tabcolsep=0.1pt
    \caption{Results of ResNet-56 on CIFAR-10}
    
    \label{cifar_resnet}
    \scalebox{1}{
        \centering
        \begin{tabular}{l|c|c|c|c|c}
            \hline
            Method         &PT?   & Acc.(\%) & $\uparrow \downarrow$(\%)& FLOPs (PR) & Params (PR)  \\
            \hline
            Baseline &N & 93.26 &0.00$\downarrow$  &125.49M (0.0\%) &0.85M (0.0\%)    \\
            \hdashline
            SNACS \cite{snacs}&Y  & 93.38&0.12$\uparrow$  &79.20M (36.9\%) &-    \\
            GAL-0.6 \cite{DBLP:conf/cvpr/LinJYZCYHD19}&Y  & 92.98&0.28$\downarrow$  &78.30M (37.6\%) &0.75M (11.8\%)    \\	
            {FilterSketch \cite{DBLP:journals/tnn/LinCLYTLTJ22}}&Y  & 93.19 &0.07$\downarrow$  &73.36M (41.5\%) &0.50M (41.2\%)    \\
            PSO \cite{pso}&Y  & 93.48&0.22$\uparrow$  &63.40M (49.5\%) &0.47M (44.7\%)    \\
            DECORE-200 \cite{DBLP:conf/cvpr/AlwaniWM22}&Y  & 93.26&0.00$\downarrow$  &62.93M (49.9\%) &0.43M (49.0\%)    \\
            HRank \cite{DBLP:conf/cvpr/LinJWZZ0020}&Y  & 93.17 &0.09$\downarrow$ &62.72M (50.0\%) &0.49M (42.4\%)    \\
            DCPH \cite{dcph}&N  & 92.96 &0.30$\downarrow$ &61.40M (51.1\%) &0.41M (51.8\%)    \\
            LRPET ($P$=0.55)&N & 93.07$\pm$0.33 &0.19$\downarrow$  & {61.20M (51.2\%) }
            &\textbf{0.41M (51.8\%)}\\
            LRPET ($P$=0.55)&Y & 93.43$\pm$0.16  &0.17$\uparrow$ & {61.20M  (51.2\%) }
            &\textbf{0.41M (51.8\%)}\\
            \textbf{LRPET$\star$($\bm{P}$=0.55)}&Y & {93.68$\pm$0.17} &\textbf{0.42$\uparrow$}  & {61.20M (51.2\%) }
            &\textbf{0.41M (51.8\%)}\\
            {CC$\dagger$ ($C$=0.5)} \cite{DBLP:conf/cvpr/LiLLYWC0M0J21}&Y & {93.53$\pm$0.11}&0.27$\uparrow$ & {60.99M (51.4\%)} &{0.45M (47.0\%)}\\
            {CC ($C$=0.5)} \cite{DBLP:conf/cvpr/LiLLYWC0M0J21}&Y & {93.64} &0.31$\uparrow$& \textbf{60.00M (52.0\%)} &{0.44M (48.2\%)}\\
            \hdashline
            {ESNB \cite{DBLP:journals/tnn/ZhouYY22}}&Y  & 93.75 &0.62$\downarrow$ & {59.48M (52.6\%)} &\textbf{0.37M (56.4\%)}\\
            FTWT$_{J}$ \cite{DBLP:conf/cvpr/ElkerdawyE0R22} &Y  & 92.28 &1.38$\downarrow$ &57.73M (54.0\%) &- \\
            SFP \cite{DBLP:conf/ijcai/HeKDFY18}&N  & 92.62$\pm$0.13 &1.33$\downarrow$ & {56.69M (54.8\%)} &-\\
            {LRPET (${P}$=0.57)}&N & {92.99$\pm$0.21} &0.27$\downarrow$  & {56.06M (55.3\%) } &{0.39M (53.5\%)}\\
            \textbf{LRPET ($\bm{P}$=0.57)}&Y & {93.27$\pm$0.14} &\textbf{0.01$\uparrow$}  & {56.06M (55.3\%) } &{0.39M (53.5\%)}\\
            {EDP \cite{DBLP:journals/tnn/RuanLYLHLM21}}&Y  & 93.61 &0.00$\downarrow$ & \textbf{53.08M (57.7\%)} &0.39M (54.1\%)\\
            \hdashline
            GAL-0.8 \cite{DBLP:conf/cvpr/LinJYZCYHD19}&Y  & 90.36 &2.90$\downarrow$ &49.99M (60.2\%) &0.29M (65.9\%)    \\
            FTWT$_{D}$ \cite{DBLP:conf/cvpr/ElkerdawyE0R22} &Y  & 92.63 &1.03$\downarrow$ &42.66M (66.0\%) &- \\
            {LRPET (${P}$=0.70)}&N & {92.13$\pm$0.17 }  &1.13$\downarrow$& \textbf{38.57M (69.2\%)} &\textbf{0.27M (67.4\%)}\\
            \textbf{LRPET ($\bm{P}$=0.70)}&Y & {92.71$\pm$0.14 }  &\textbf{0.55}$\downarrow$& \textbf{38.57M (69.2\%)} &\textbf{0.27M (67.4\%)}\\
            \hdashline
            
            HRank \cite{DBLP:conf/cvpr/LinJWZZ0020}&N  & 90.72 &2.54$\downarrow$ &32.52M (74.1\%) &0.27M (68.2\%)    \\
            {FilterSketch \cite{DBLP:journals/tnn/LinCLYTLTJ22}}&Y  & 91.20&2.06$\downarrow$  &32.47M (74.4\%) &0.24M (71.8\%)    \\
            LRPET ($P$=0.80)&N & {91.13$\pm$0.59} &2.13$\downarrow$& 26.23M (79.1\%) &\textbf{0.17M (79.0\%)}\\
            
            \textbf{LRPET-S}&N & {91.24$\pm$0.23} &\textbf{2.02}$\downarrow$&\textbf{24.13M (80.8$\%$)}&{0.18M (78.9$\%$)}\\	
            
            \hline
    \end{tabular}}
\end{table}

\begin{table}[t]
    \footnotesize
    \centering
    \tabcolsep=0.5pt 
    \caption{Results of ResNet-110 on CIFAR-10}
    \label{cifar_resnet110}
    \scalebox{1}{
    \centering
    \begin{tabular}{l|c|c|c|c|c}
		\hline
		Method         &PT?   & \tabincell{c}{Acc.(\%)} & $\uparrow \downarrow$(\%) & FLOPs (PR) & Params (PR)  \\
			
		\hline
		Baseline &N & 93.50 &0.00$\downarrow$ &252.89M(0.0\%) &1.72M(0.0\%) \\
		\hdashline
		GAL-0.5 \cite{DBLP:conf/cvpr/LinJYZCYHD19}&Y & 92.55&0.95$\downarrow$ &130.20M (48.5\%)&0.95M (44.8\%)\\
		DCPH \cite{dcph}&N & 93.54&0.04$\uparrow$ & 123.50M (51.2\%) &0.82M (52.3\%)\\
		HRank \cite{DBLP:conf/cvpr/LinJWZZ0020}&N & {93.36} &0.14$\downarrow$& 105.70M (58.2\%) & 0.70M (59.3\%)          \\
		DECORE-300 \cite{DBLP:conf/cvpr/AlwaniWM22}&Y  & 93.50  &0.00$\downarrow$&96.66M (61.8\%) &\textbf{0.61M (64.8\%)}    \\
   	SFP \cite{DBLP:conf/ijcai/HeKDFY18}&N  & 92.90$\pm$0.23 &0.30$\downarrow$ & 94.03M (62.8\%) &-\\
		\textbf{LRPET ($\bm{P}$=0.65)}&N & {93.61$\pm$0.19} &\textbf{0.11}$\uparrow$&93.78M (62.9\%)&{0.65M (61.8\%)}\\
        {FilterSketch \cite{DBLP:journals/tnn/LinCLYTLTJ22}}&Y  & 93.44 &0.06$\downarrow$  &\textbf{92.84M (63.3\%)} &0.69M (59.9\%)    \\	
		\hline
    \end{tabular}}
\end{table}

\textbf{ResNets.} We report the classification accuracy, FLOPs \cite{patterson2016computer}, and parameter sizes accompanied with the corresponding pruning rate (PR) of 
ResNet-18, ResNet-50, ResNet-56 and ResNet-110 in Table \ref{cifar_resnet18}, \ref{cifar_resnet50}, \ref{cifar_resnet} and \ref{cifar_resnet110}, respectively. Here FLOPs refer to the number of floating point operations required by the model to process an image. It must be noted that for the sake of fairness, our training configuration is the same as \cite{edropout} in Table \ref{cifar_resnet18} and \ref{cifar_resnet50}. We change the rank pruning ratio $P$ of LRPET to get results with different FLOPs and parameter sizes. The dashed lines separate results with the same order of FLOPs and the best results are marked in bold. 
Note that the data augmentation used by CC \cite{DBLP:conf/cvpr/LiLLYWC0M0J21} is not the same as the standard one, i.e., each image side is zero-padded with 8 pixels. For {\color{red}a} fair comparison, we also use the same configuration as CC \cite{DBLP:conf/cvpr/LiLLYWC0M0J21} for training, denoted by LRPET$\star$. Since CC \cite{DBLP:conf/cvpr/LiLLYWC0M0J21} only provides the result of one trial, for more thorough comparison, we also reproduce their experiment with the official provided code for {three times}, denoted by CC$\dagger$. 

We can see that, compared with other methods, LRPET achieves higher accuracy increment or lower accuracy reduction with less or comparable FLOPs and parameters for both sizes of ResNets.
Note that most compared methods only report the result of one trial while we report the average result of three trials.  For a fair comparison, we re-conduct experiments three times for all comparison methods that have provided publicly available code and detailed configurations.

Compared with the state-of-the-art pruning methods, the success of LRPET reveals that the overlooked low-rankness can outperform the pruning methods with proper design. Note that FTWT is a dynamic pruning method, which is expected to outperform the static compression method the pruned network structure adaptively changes according to different samples. However, LRPET still outperforms it at different computation budgets. The results of ResNet-110 show that LRPET without pre-training even outperforms the pruning method DECORE with pre-trained model.
Although LRPET with default data augmentation performs a little poorer than CC, LRPET outperforms CC when using the same data augmentation, and the superiority is more significant in the case of multiple trials. It is worth noting that CC is the combination of low-rank decomposition and pruning while LRPET only utilizes low-rankness. Besides, LRPET trains the network only for 200 epochs, but CC needs 300 epochs.

\begin{figure}[t]
	\centering
	\includegraphics[width=0.45\textwidth]{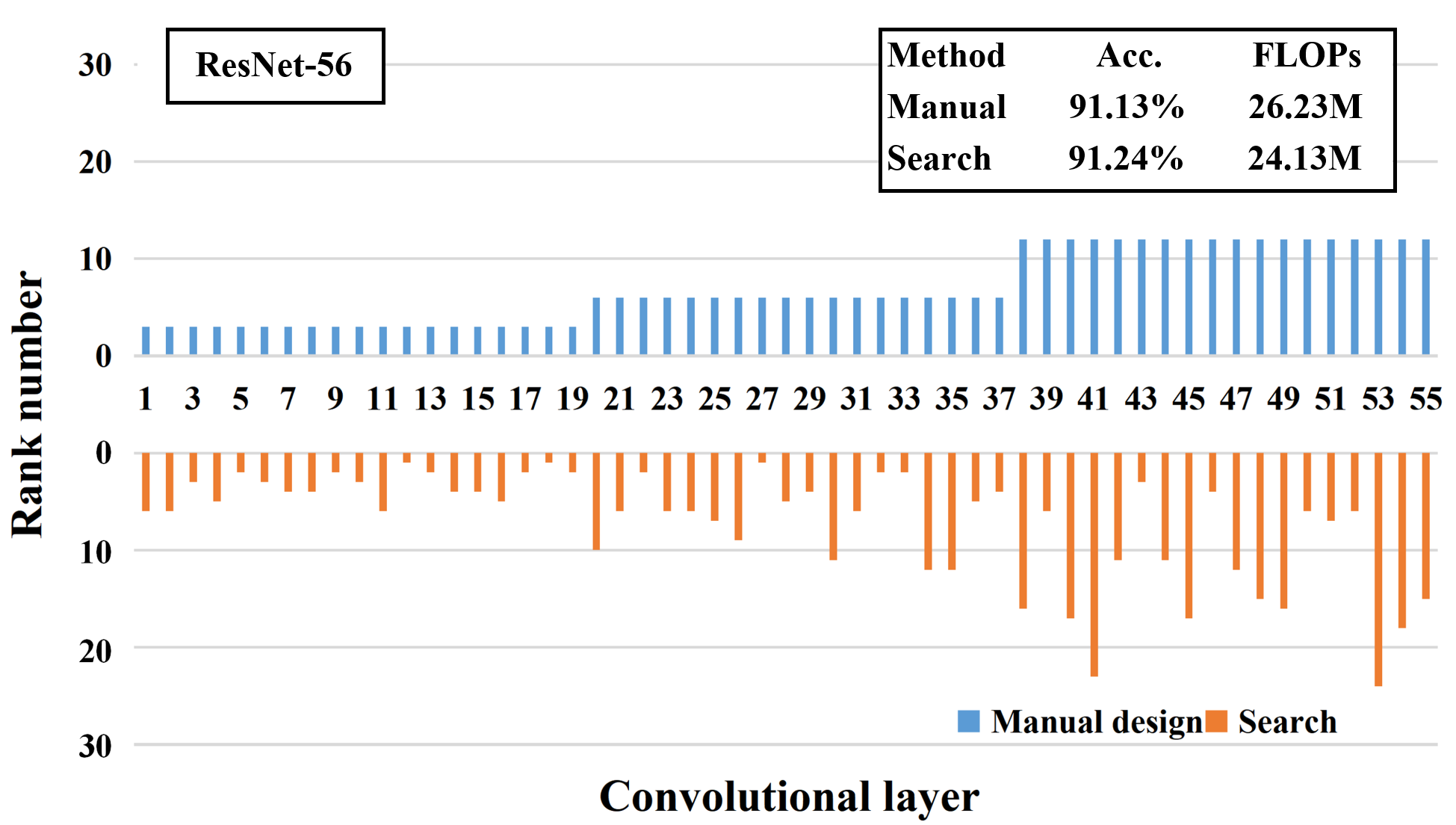}
	\caption{Rank distribution comparison between manual fixed layer rank ratio setting and searched layer rank setting for ResNet-56 on CIFAR-10.}
	\label{fig:search}
\end{figure}

We conduct a Bayesian optimization experiment on ResNet-56 to show that our result can be further improved by searching hyperparameters. From Table \ref{cifar_resnet}, LRPET-S achieves a higher reduction of FLOPs (80.8$\%$ vs. 79.1$\%$) with better performance (91.24$\%$ vs. 91.13$\%$) than LRPET.  We also display the rank number of different layers for the two methods in Fig. \ref{fig:search}. We can see that LRPET-S adaptively sets the rank pruning ratio for different layers, which suggests that different layers have different sensitivity to network compression.

Note that for the shortcut between residual blocks with different channel numbers, TRP \cite{DBLP:conf/ijcai/XuL0WWQCLX20} and SVD training \cite{DBLP:conf/cvpr/YangTW0HLLC20} use $1\times1$ convolution (shortcut B) while the other methods adopt the padding strategy (shortcut A). To compare with them, following TRP, 
we apply LRPET to ResNet 20/56 with shortcut B (denoted by ResNet-20/56-B) for 164 epochs. 
To show that our superiority does not simply come from the weight decay setting, we re-implement TRP for the weight decay of 5e-4 and quote the result for the weight decay of 1e-4 from their paper. 
Following the official code of TRP, we also fine-tune the networks for four epochs in our re-implementation. Since the speeding up of TRP varies in different trials, we report the one with the best accuracy from three trials for TRP re-implementation. 
Note that SVD training utilizes pre-training and fine-tuning, which results in three times of total training epochs. For a fair comparison, we also experiment LRPET with pre-trained models, denoted by  LRPET\#. The settings are the same as the above experiments except that the training epoch number of LRPET is 300.
The results are shown in Table \ref{cifar_resnet_trp}. The performance gain of LRPET over the optimal result of TRP is significant. We should emphasize that TRP needs to use SVD for every iteration to guarantee good performance, which takes much more time than LRPET. 
Besides, the four epochs of fine-tuning effectively improve TRP's performance while LRPET does not need fine-tuning to achieve good performance. SVD training is better than LRPET trained with 164 epochs, but the superiority comes from more training epochs. When training with a fair number of epochs, LRPET outperforms SVD training, which shows that preserving the model capacity improves the performance.
\begin{table}[t]
	\footnotesize
	\centering
	\tabcolsep=5pt
	\caption{Results of ResNet-20/56-B on CIFAR-10
	}
	\label{cifar_resnet_trp}
	\scalebox{1}{
		\centering
		\begin{tabular}{l|c|c}
			\hline
			Method         & Acc. (\%) & Speed up   \\
			\hline
			
			ResNet-20-B (Baseline, wd=1e-4) & 91.74  &1.00$\times $    \\
			ResNet-20-B (Baseline, wd=5e-4) & 92.40  &1.00$\times $    \\
			ResNet-20-B (TRP1, wd=1e-4)& 90.12& 1.97$\times $\\
			ResNet-20-B (TRP1+Nu, wd=1e-4) & 90.50  &2.17$\times $    \\
            ResNet-20-B (TRP1, wd=5e-4)& {86.00}& {1.30$\times $}\\
			ResNet-20-B (TRP1+Nu, wd=5e-4) & {85.52}  &{1.27$\times $}    \\
			{ResNet-20-B (LRPET, wd=5e-4, ${P}$=0.60)} & {90.62$\pm$0.24}  &\textbf{2.34$\times $}   \\
            {ResNet-20-B (SVD training \cite{DBLP:conf/cvpr/YangTW0HLLC20}, wd=1e-4)}  & 90.97  &2.20$\times $    \\
            {{ResNet-20-B (LRPET\#, wd=5e-4, $\bm{P}$=0.60)}} & {\textbf{91.56$\pm$0.10}}  &\textbf{2.34$\times $}   \\
			\hline
			ResNet-56-B (Baseline, wd=1e-4) & 93.14 &1.00$\times $    \\
			ResNet-56-B (Baseline, wd=5e-4) & 93.52 &1.00$\times $    \\
			ResNet-56-B (TRP, wd=1e-4) &92.77 &2.31$\times $ \\
            ResNet-56-B (TRP, wd=5e-4) &{92.72} &{1.57$\times $ }\\
            ResNet-56-B (TRP+Nu, wd=5e-4) &{92.16} &{1.50$\times $ }\\
			{ResNet-56-B (LRPET, wd=5e-4, $\bm{P}$=0.65)} & \textbf{92.89$\pm$0.08}  &\textbf{2.50$\times $}  \\
            \hdashline
                {ResNet-56-B (SVD training \cite{DBLP:conf/cvpr/YangTW0HLLC20}, wd=1e-4)}  & 93.67  &2.70$\times $    \\
            {ResNet-56-B (LRPET\#, wd=5e-4, $\bm{P}$=0.69)} & {\textbf{93.69$\pm$0.18}}  &{{3.01$\times $}}  \\
			ResNet-56-B (TRP1+Nu, wd=1e-4) & 91.85  & \textbf{4.48$\times$}  \\
			{ResNet-56-B (LRPET, wd=5e-4, ${P}$=0.81)} & {92.05$\pm$0.19}  &{4.44$\times $}  \\
			\hline
	\end{tabular}}
\end{table}

\textbf{GoogLeNet}. The results of GoogLeNet on CIFAR-10 are shown in Table \ref{cifar_google}. LRPET achieves the highest accuracy with the smallest FLOPs. It even outperforms the baseline when cutting about half of the FLOPs and parameters. Without pre-training, LRPET outperforms pruning methods with pre-trained models (DECORE, CC, and HRank). This again demonstrates the potential of low-rankness for network compression. Compare with the pruning method GAL which also trains from scratch, the performance gain of LRPET is significant (more than 1\%).

\begin{table}[ht]
	\footnotesize
    \centering
    \tabcolsep=1pt
    \caption{Results of GoogLeNet on CIFAR-10}
    \label{cifar_google}
    \scalebox{1}{
        \centering
        \begin{tabular}{l|c|c|c|c|c}
            \hline
            Method         &PT?   & \tabincell{c}{Acc. (\%)} & $\uparrow \downarrow$(\%)  & FLOPs (PR) & Params (PR)  \\
            \hline
            Baseline &N & 95.05 &0.00$\downarrow$ &1.52B(0.0\%) &6.15M(0.0\%)    \\
            \hdashline
            DECORE-500 \cite{DBLP:conf/cvpr/AlwaniWM22}&Y  & 95.20 &0.15$\uparrow$ &1.22B (19.8\%) &4.73M (23.0\%)    \\
            GAL-0.05 \cite{DBLP:conf/cvpr/LinJYZCYHD19}&Y  & 93.93 &1.12$\downarrow$ &0.94B (38.2\%) &3.12M (49.3\%)    \\
            CC ($C$=0.5) \cite{DBLP:conf/cvpr/LiLLYWC0M0J21}&Y  & 95.18  &0.13$\uparrow$ &0.76B (50.0\%) &\textbf{2.83M (54.0\%) }   \\
            
            {LRPET (${P}$=0.60)}&N & {95.25$\pm$0.23} &\textbf{0.20}$\uparrow$  & \textbf{0.74B (51.4\%)}  &{3.03M (50.8\%)}\\
            \hdashline
            HRank \cite{DBLP:conf/cvpr/LinJWZZ0020}&Y  & 94.53  &0.52$\downarrow$&0.69B(54.6\%) &2.74M (55.4\%)    \\
            CC ($C$=0.6) \cite{DBLP:conf/cvpr/LiLLYWC0M0J21} &Y & 94.88  &0.17$\downarrow$&0.61B (59.9\%) &\textbf{2.26M (63.3\%)}   \\
            {FilterSketch \cite{DBLP:journals/tnn/LinCLYTLTJ22}}&Y  & 94.88 &0.17$\downarrow$  &0.59B (61.1\%) &2.61M (57.6\%)    \\
            {LRPET (${P}$=0.68)}&N & {94.97$\pm$0.10} &\textbf{0.08}$\downarrow$ & \textbf{0.59B (61.1\%)} &{2.41M (60.8\%)}\\

            \hline
    \end{tabular}}
\end{table}

\textbf{VGG-16.} {Since the channel number of each layer of VGG-16 varies significantly, to avoid too few channels, we set the smallest ranks of the first two layers to 2/3 of the original matrix ranks and the minimum ranks of the other convolutional layers to the minimum rank of the second convolutional layer.}

Table \ref{cifar_vgg16} displays the results of VGG-16 on CIFAR-10.
Although LRPET without pre-training is a little poorer than methods with pre-trained models for some computation budgets, the performance of LRPET is significantly improved when using pre-trained model for initialization and finally outperforms the other pre-trained methods. Using pre-trained model in practice usually means nearly twice of total training time. We should emphasize that pre-trained model is necessary for these compared methods but is optional for LRPET. 
Without pre-training, LRPET still works well and is significantly better than the training from scratch method GAL.

\begin{table}[ht]
	\footnotesize
    \centering
    \tabcolsep=0.5pt 
    \caption{Results of VGG-16 on CIFAR-10}
    \setlength{\tabcolsep}{0.3pt}
    \label{cifar_vgg16}
    \scalebox{1}{
        \centering
        \begin{tabular}{l|c|c|c|c|c}
            \hline
            Method     &PT?   & \tabincell{c}{Acc. (\%)} & $\uparrow \downarrow$(\%) & FLOPs (PR) & Params (PR)  \\
            
            \hline

            Baseline &N & 93.70 &0.00$\downarrow$ &313.2M (0.0\%) &14.72M (0.0\%)    \\
            \hdashline
            DCPH \cite{dcph}&N  & 93.19 &0.51$\downarrow$ &195.00M (37.8\%) &5.50M (62.7\%)    \\
            GAL-0.05 \cite{DBLP:conf/cvpr/LinJYZCYHD19}&Y  & 92.03 &1.93$\downarrow$ &189.49M (39.6\%) &3.36M (77.6\%)    \\
            GAL-0.1 \cite{DBLP:conf/cvpr/LinJYZCYHD19}&Y  & 90.78&3.18$\downarrow$  &171.89M (45.2\%) &2.67M (82.2\%)    \\
            HRank \cite{DBLP:conf/cvpr/LinJWZZ0020}&Y  & 93.43&0.53$\downarrow$  &145.61M (53.5\%) &2.51M (82.9\%)  \\
            LRPET ($P$=0.63)  &N & 93.49$\pm$0.14 &0.21$\downarrow$ & {144.10M (54.0\%)}  &{6.12M (58.5\%)} \\
            {LRPET ($\bm{P}$=0.63)}  &Y & {93.86$\pm$0.25}&\textbf{0.16}$\uparrow$  & \textbf{144.10M (54.0\%)}  &{6.12M (58.5\%)} \\
            {FTWT$_{D}$  \cite{DBLP:conf/cvpr/ElkerdawyE0R22}}&{Y}  & {93.73}&0.09$\uparrow$  &{137.81M (56.0\%)} &{-} \\
            {PSO  \cite{pso}}&{Y}  & {93.73}&0.15$\uparrow$  &{137.80M (56.0\%)} &{\textbf{2.31M (84.6\%)}} \\
            {LRPET ($P$=0.65) } &{N} & {93.46$\pm$0.10}  &0.24$\downarrow$& {\textbf{137.78M (56.0\%)}}  &{5.79M (60.6\%)}  \\
            {{LRPET (${P}$=0.65)}}  &{Y} &{93.76$\pm$0.12}&0.06$\uparrow$ & {\textbf{137.78M (56.0\%)}}  &{5.79M (60.6\%)} \\
            \hdashline
            DECORE-200 \cite{DBLP:conf/cvpr/AlwaniWM22}&Y  & 93.56&0.40$\downarrow$  &110.51M (64.8\%) &\textbf{1.66M (89.0\%)}    \\
            FTWT$_{J}$  \cite{DBLP:conf/cvpr/ElkerdawyE0R22}&Y  & 93.55  &0.27$\downarrow$&109.62M (65.0\%) &- \\
            HRank \cite{DBLP:conf/cvpr/LinJWZZ0020}&Y  & 92.34  &1.62$\downarrow$&108.61M (65.3\%) &2.64M (82.4\%)    \\
            LRPET ($P$=0.77)  &N & {93.27$\pm$0.17} &0.43$\downarrow$& \textbf{107.56M (65.7\%)}  &{3.83M (74.0\%)}  \\
            {LRPET (${P}$=0.77)}  &Y &{93.70$\pm$0.14} &\textbf{0.00}$\downarrow$& \textbf{107.56M (65.7\%)}  &{3.83M (74.0\%)}  \\
            \hdashline
            SNACS  \cite{snacs} &Y  & 91.06 &2.64$\downarrow$ &103.83M (67.9\%) &- \\
            FTWT$_{D}$  \cite{DBLP:conf/cvpr/ElkerdawyE0R22} &Y  & 93.19 &0.63$\downarrow$ &84.56M (73.0\%) &- \\
            FTWT$_{J}$ \cite{DBLP:conf/cvpr/ElkerdawyE0R22} &Y  & 92.65  &1.17$\downarrow$&81.43M (74.0\%) &- \\
            LRPET ($P$=0.92)&N & 92.93$\pm$0.11 &0.77$\downarrow$& \textbf{81.29M (74.1\%)} &\textbf{1.62M (88.9\%)}\\
            {LRPET (${P}$=0.92)}&Y & {93.52$\pm$0.05}&\textbf{0.18}$\downarrow$& \textbf{81.29M (74.1\%)}  &\textbf{1.62M (88.9\%)}\\
            \hline
    \end{tabular}}
\end{table}

\textbf{FBNet.} Our method has achieved good compression effect in the above convolution models, which are classical but also redundant. To show the advancedness of LRPET, we further explore the compression effect of LRPET on compact models. We chose FBNet-B \cite{wu2019fbnet} as the experimental object since it is a compact model obtained from neural architecture search methods. As FBNet was originally designed to be used on the ImageNet, we modified FBNet-B to be more compatible with images of the size $32\times32$.
For the compared methods, we conduct experiments with publicly available codes. Table \ref{cifar_fbnet} displays the results of FBNet on CIFAR-10. LRPET outperforms the other methods. LRPET achieves almost the same accuracy as baseline when pruning 25.0\% of the parameters. Since FBNet-B is a more compact network, LRPET cannot compress FBNet-B as much as the previous networks without too much performance degeneration. However, LRPET still outperforms the compared methods.

\begin{table}[ht]
	\footnotesize
    \centering
    \tabcolsep=0.5pt
    \caption{Results of FBNet on CIFAR-10}
    \setlength{\tabcolsep}{0.3pt}
    \label{cifar_fbnet}
    \scalebox{1}{
        \centering
        \begin{tabular}{l|c|c|c|c|c}
            \hline
            Method   & PT? &\tabincell{c}{Acc. (\%)} & $\uparrow \downarrow$(\%) & FLOPs (PR) & Params (PR)  \\
            
            \hline
            Baseline & N & 94.73 &0.00$\downarrow$ &281.4M (0.0\%) &2.85M (0.0\%)    \\
            \hdashline
            HRank \cite{DBLP:conf/cvpr/LinJWZZ0020}  & N &  94.26$\pm$0.21 &0.47$\downarrow$ &220.07M (21.8\%) &2.29M (19.7\%)    \\
            FilterSketch \cite{DBLP:journals/tnn/LinCLYTLTJ22}  &N &  92.41$\pm$0.27 &2.32$\downarrow$  &220.65M (21.6\%) &2.26M (20.8\%)    \\
            SFP \cite{DBLP:conf/ijcai/HeKDFY18}  & N & 93.11$\pm$0.19&1.62$\downarrow$  &220.66M (21.7\%) & -    \\
            LRPET ($P$=0.40)   & N & 94.66$\pm$0.08 &\textbf{0.07$\downarrow$} & \textbf{218.11M(22.5\%)}  & \textbf{2.14M(25.0\%)} \\
            \hline
    \end{tabular}}
\end{table}

\textbf{GhostNet.} {
We also conduct experiments for the manually designed leightweight network GhostNet \cite{DBLP:conf/cvpr/HanW0GXX20}
	The results of Ghost-ResNet56 \cite{DBLP:conf/cvpr/HanW0GXX20} on CIFAR-10 are shown in Table \ref{cifar_ghostnet}. Compared with other relevant methods, LRPET achieves the highest accuracy with the smallest FLOPs, which again verifies that LRPET can further compress advanced compact networks. }

\begin{table}[ht]
	\footnotesize
    \centering
    \tabcolsep=0.5pt 
    \caption{Results of Ghost-ResNet56 on CIFAR-10}
    \setlength{\tabcolsep}{0.3pt}
    \label{cifar_ghostnet}
    \scalebox{1}{
        \centering
        \begin{tabular}{l|c|c|c|c|c}
            \hline
            Method   & PT? &\tabincell{c}{Acc. (\%)} & $\uparrow \downarrow$(\%) & FLOPs (PR) & Params (PR)  \\
            
            \hline

            Baseline & N & 91.88 &0.00$\downarrow$ &67.80M (0.0\%) &0.44M (0.0\%)    \\
            \hdashline
            HRank \cite{DBLP:conf/cvpr/LinJWZZ0020}  &  N & 91.07$\pm$0.17 &0.81$\downarrow$ &45.56M (32.8\%) &0.29M (34.1\%)    \\
            FilterSketch \cite{DBLP:journals/tnn/LinCLYTLTJ22}  & N &  88.42$\pm$0.32 &3.46$\downarrow$  &42.58M (36.9\%) &0.27M (39.0\%)    \\
            SFP \cite{DBLP:conf/ijcai/HeKDFY18}  &  N & 90.92$\pm$0.22 &0.96$\downarrow$  &42.42M (37.4\%) & -    \\
            FTWT$_{D}$ \cite{DBLP:conf/cvpr/ElkerdawyE0R22}  & N &  88.42$\pm$0.27 &3.46$\downarrow$  &42.23M (37.7\%) & -    \\
            LRPET ($P$=0.40)   &  N &91.63$\pm$0.18 &\textbf{0.25$\downarrow$} & \textbf{41.64M(38.6\%)}  & 0.28M(36.8\%) \\
            \hline
    \end{tabular}}
\end{table}

\textbf{Transformer-based Model.} To demonstrate that our low-rank decomposition method has good transferability, we also apply it in Transformer-based models. Due to issues with computing resources and training cycles, we trained the ViT\cite{vit2020image} from scratch on small-size datasets, and the specific experimental configuration was the same as\cite{smallvit}. The results are shown in Table\ref{transformer}. Compared with the classic Token pruning method, our method can effectively reduce the amount of model calculations and parameters while having less impact on model accuracy.

\begin{table}[ht]
	\footnotesize
	\centering
	\tabcolsep=2.0pt 
	\caption{Results of ViT on CIFAR}
	\label{transformer}
	\scalebox{1}{
		\centering
		\begin{tabular}{l|c|c|c|c}
			\hline
			Method & FLOPs (PR) & Params (PR) &CIFAR10 &CIFAR100  \\
			
			\hline
			Baseline  &174.25M(0.0\%) &2.69M(0.0\%) &93.54 &72.49    \\
			\hdashline
			EViT\cite{evit2022not} &  128.78M (26.1\%)&2.69M (0.0\%)&92.67& 71.83    \\
			LRPET &  126.42M (27.4\%)&1.97M (26.8\%)&92.96& 72.08    \\
			\hline
	\end{tabular}}
\end{table}

\subsubsection{Results on ImageNet} 
To show the generalization ability of LRPET, experiments on large-scale dataset ImageNet with ResNet 18/34/50 are conducted and the results are shown in Table \ref{imagenet}. 
We can see that LRPET also outperforms the other state-of-the-art low-rank decomposition and pruning methods. It achieves the highest top-1 and top-5 accuracy with the smallest FLOPs for all three networks. At most FLOPs orders, LRPET without pre-training performs better than methods either with or without pre-training. When pre-trained models are utilized, the top-1 accuracy of LRPET is improved with more than 1\% and achieves the best performance for all the cases. Specifically, compared with baseline, LRPET for ResNet-50 shows only 0.24\% decrease in top-1 accuracy when the FLOPs reduction rate is 53.6\%. From the results of ResNet-50, we empirically find that the superiority of LRPET is more prominent for larger FLOPs reduction rate. When $P$=0.80, the performance gain of LRPET without pre-training over the second best method (pre-trained pruning method DECORE) is 1.26\%, which is very significant for large-scale dataset. 
Compare with the baseline of ResNet-18, LRPET ($P$=0.80) achieves higher accuracy with only nearly half of the FLOPs. For the hyper-parameter search experiment of LRPET-S, although rank pruning ratios are searched on CIFAR-10, they are transferable to ImageNet. LRPET-S with and without pre-training achieve higher accuracy with smaller FLOPs than all the compared methods of the same FLOPs order.

\begin{table*}[t]
	\footnotesize

    \centering
    \caption{Results of ResNets on ImageNet }
    \label{imagenet}
    \scalebox{1}{
        \centering
        \begin{tabular}{l|l|c|c|c|c|c|c|c}
            \hline
            Model   &Method     &PT? & {Top-1 Acc. (\%)} &$\uparrow \downarrow$(\%) & {Top-5  Acc. (\%)} &$\uparrow \downarrow$(\%) & FLOPs (PR) & Params (PR)  \\
            \hline
            \multirow{6}{*}{ResNet-18}
            &Baseline &N & 69.76 &0.00$\downarrow$ & 89.08 &0.00$\downarrow$  &1.82B (0.0\%) &11.69M (0.0\%)    \\
            \cdashline{2-9}
            &SFP \cite{DBLP:conf/ijcai/HeKDFY18} &N & 67.10&3.18$\downarrow$ & 87.78 &1.85$\downarrow$ &1.06B (41.8\%) &-   \\
            &{EPruner-0.73 \cite{DBLP:journals/tnn/LinJLWWHY22}} &Y & 67.31 &2.35$\downarrow$& 87.70 &1.38$\downarrow$&1.02B (43.8\%) &6.05M (48.2\%)    \\
            &TRP \cite{DBLP:conf/ijcai/XuL0WWQCLX20} &N & 65.46 &3.64$\downarrow$& 86.48  &2.46$\downarrow$&1.01B (44.5\%) &-   \\
            
            &FTWT ($r$=0.91) \cite{DBLP:conf/cvpr/ElkerdawyE0R22} &Y & 67.49 &2.27$\downarrow$ & - &- &0.88B (51.6\%) &-    \\
            &{LRPET (${P}$=0.58)} &N & {67.87} &\textbf{1.89}$\downarrow$&{88.04}&\textbf{1.04}$\downarrow$ & \textbf{0.86B (52.5\%)} &\textbf{5.81M (50.2\%)}\\
            
            \hline
            \multirow{8}{*}{ResNet-34}
            
            &Baseline &N  & 73.31&0.00$\downarrow$  & 91.42&0.00$\downarrow$  &3.66B (0.0\%) &21.78M (0.0\%)    \\
            \cdashline{2-9}
            &EDropout \cite{edropout} &N & 71.82 &1.60$\downarrow$& 90.96 &\textbf{0.30$\downarrow$} &- &10.52M (51.7\%)   \\
            &SFP \cite{DBLP:conf/ijcai/HeKDFY18} &N & 71.83 &2.09$\downarrow$& 90.33 &1.29$\downarrow$ &2.16B (41.1\%) &-   \\
            &{EDP \cite{DBLP:journals/tnn/RuanLYLHLM21}} &Y & 72.33 &1.13$\downarrow$& 90.88 &0.51$\downarrow$&2.02B (44.9\%) &11.87M (45.5\%)    \\
            &{EPruner-0.75 \cite{DBLP:journals/tnn/LinJLWWHY22}} &Y & 70.95 &2.23$\downarrow$& 89.97 &1.48$\downarrow$&1.85B (49.6\%) &10.24M (53.2\%)    \\

            &FTWT ($r$=0.92) \cite{DBLP:conf/cvpr/ElkerdawyE0R22} &Y & 71.71  &1.59$\downarrow$& - &- &1.75B (52.2\%) &-    \\
            
            &{LRPET ($P$=0.58)}  &N & {71.84} &1.47$\downarrow$&90.48 &0.94$\downarrow$& \textbf{1.71B (53.1\%)} & \textbf{10.51M (51.7\%)}\\
            &{LRPET (${P}$=0.58)} &Y & {72.95} &\textbf{0.36}$\downarrow$&{90.98} &0.44$\downarrow$& \textbf{1.71B (53.1\%)} &\textbf{10.51M (51.7\%)}\\
            \hline
            \multirow{27}{*}{ResNet-50}
            
            &Baseline &N & 76.15 &0.00$\downarrow$  & 92.87&0.00$\downarrow$ &4.09B (0.0\%) &25.50M (0.0\%) \\
            \cdashline{2-9}
            &{FilterSketch-0.7 \cite{DBLP:journals/tnn/LinCLYTLTJ22}} &Y & 75.22&0.91$\downarrow$ & 92.41 &0.45$\downarrow$ &2.64B (35.5\%) &16.95M (33.5\%)   \\
            &{SNACS \cite{snacs}} &Y & 72.60&3.55$\downarrow$ &- &- &2.38B (41.8\%) &-   \\
            &SFP \cite{DBLP:conf/ijcai/HeKDFY18} &N & 74.61&1.54$\downarrow$& 92.06&0.81$\downarrow$  &2.38B (41.8\%) &-   \\
            &DECORE-6 \cite{DBLP:conf/cvpr/AlwaniWM22} &Y & 74.58 &1.57$\downarrow$& 92.18 &0.69$\downarrow$ &2.36B (42.3\%) &14.10M (44.71\%)   \\
            &GAL-0.5 \cite{DBLP:conf/cvpr/LinJYZCYHD19} &Y & 71.95 &4.20$\downarrow$& 90.94&1.93$\downarrow$ &2.33B (43.0\%)  &21.20M (16.9\%)\\
            
            &HRank \cite{DBLP:conf/cvpr/LinJWZZ0020} &Y & {74.98} &1.17$\downarrow$    &{92.33} &0.54$\downarrow$      & 2.30B (43.9\%)    & 16.15M (36.7\%)   \\
            &TRP+Nu \cite{DBLP:conf/ijcai/XuL0WWQCLX20} &N & 74.06 &2.09$\downarrow$   &92.07  &0.63$\downarrow$    & 2.27B (44.5\%)   & -   \\

            &{FilterSketch-0.6 \cite{DBLP:journals/tnn/LinCLYTLTJ22}} &Y & 74.68&1.45$\downarrow$ & 92.17 &0.69$\downarrow$ &2.23B (45.5\%) &14.53M (43.0\%)   \\
            &{ESNB \cite{DBLP:journals/tnn/ZhouYY22} }&Y & 76.13&1.14$\downarrow$ & 93.02 &0.59$\downarrow$ &2.12B (48.1\%) &-   \\
            &{EDP \cite{DBLP:journals/tnn/RuanLYLHLM21}} &Y & 75.34 &1.17$\downarrow$& 92.43 &0.34$\downarrow$&1.94B (52.6\%) &14.30M (43.9\%)    \\
            &CC ($C$=0.5)\cite{DBLP:conf/cvpr/LiLLYWC0M0J21} &Y &{75.59}&0.56$\downarrow$ & {92.64}&0.23$\downarrow$&1.93B (52.9\%)&13.20M (48.4\%)\\
            &{EPruner-0.73 \cite{DBLP:journals/tnn/LinJLWWHY22}} &Y & 74.26 &1.75$\downarrow$& 91.88 &1.08$\downarrow$&1.92B (53.3\%) &12.70M (50.3\%)    \\
            
            &{LRPET ($P$=0.62)} &N & {74.25}&1.90$\downarrow$ &{91.93}&0.94$\downarrow$  &\textbf{1.90B (53.6\%)}&\textbf{12.89M (49.5\%)}\\
            &{LRPET (${P}$=0.62)} &Y & {75.91} &\textbf{0.24}$\downarrow$&{92.79} &\textbf{0.08$\downarrow$} &\textbf{1.90B (53.6\%)}&12.89M (49.5\%)\\
            \cdashline{2-9}
            &EDropout \cite{edropout} &N & 73.72 &2.43$\downarrow$& 91.21&1.66$\downarrow$ &-  &10.89M (57.3\%)\\
            &GAL-0.5-joint \cite{DBLP:conf/cvpr/LinJYZCYHD19} &Y & 71.80 &4.35$\downarrow$& 90.92&1.95$\downarrow$ &{1.84B (55.0\%)}  &19.31M (24.3\%)\\
            &PSO \cite{pso} &Y & 71.93 &4.22$\downarrow$& 91.26&1.61$\downarrow$ &1.71B (58.1\%) &12.20M (52.2\%)   \\
            &DECORE-5 \cite{DBLP:conf/cvpr/AlwaniWM22} &Y & 72.06 &4.09$\downarrow$& 90.82&2.05$\downarrow$ &1.60B (60.9\%) &\textbf{8.87M (65.2\%)}   \\
            &HRank \cite{DBLP:conf/cvpr/LinJWZZ0020} &N & 71.89&4.26$\downarrow$      &91.01&1.86$\downarrow$       & 1.55B (62.1\%)    & 13.77M (46.0\%)   \\
            
            &{FilterSketch-0.4 \cite{DBLP:journals/tnn/LinCLYTLTJ22}} &Y & 73.04&3.09$\downarrow$ & 91.18 &1.68$\downarrow$ &1.51B (63.1\%) &10.40M (59.2\%)   \\
            &{LRPET-S}  &N  & {72.50}&3.65$\downarrow$ &{90.99} &1.88$\downarrow$ &\textbf{1.38B (66.2\%)}&{9.18M (64.0\%)}\\
            &{LRPET-S}  &Y  & {73.72} &\textbf{2.43}$\downarrow$&{91.73} &\textbf{1.14}$\downarrow$ &\textbf{1.38B (66.2\%)}&{9.18M (64.0\%)}\\
            \cdashline{2-9}
            &DECORE-4 \cite{DBLP:conf/cvpr/AlwaniWM22} &Y & 69.71 &6.44$\downarrow$& 89.37 &3.50$\downarrow$&1.19B (70.9\%) &\textbf{6.12M (76.0\%)}   \\
            &GAL-1-joint \cite{DBLP:conf/cvpr/LinJYZCYHD19} &Y & 69.31 &6.84$\downarrow$& 89.12& 3.75$\downarrow$&1.11B (72.9\%)  &10.12M (60.3\%)\\
            &HRank \cite{DBLP:conf/cvpr/LinJWZZ0020} &N & 69.10&7.05$\downarrow$& 89.58 &3.29$\downarrow$&\textbf{0.98B (76.0\%)}  &8.27M (67.6\%)\\
            &{LRPET (${P}$=0.80)} &N & {70.97} &\textbf{5.18}$\downarrow$&{90.16} &\textbf{2.71}$\downarrow$ &\textbf{0.98B (76.0\%)}&{7.74M (70.0\%)}\\

            \hline
    \end{tabular}}
\end{table*}

\subsection{Generalization to Other Vision Tasks}

\begin{table}[ht]
	\footnotesize
	\centering
	\caption{Results for Object Detection on PASCAL VOC 2007} 
	\label{detection}
	\centering
	\begin{tabular}{c|c|c|c}
		\hline
		Model & Method & FLOPs Reduction & mAP \\
		\hline
		\multirow{2}*{Faster-RCNN} & Baseline & 0\%   & 80.78   \\
		~ & LRPET & 50\% & 80.73 \\
		\hline
	\end{tabular}
\end{table}

To further analyze the generalization of our method, we apply LRPET to the tasks of object detection and semantic segmentation.

For objection detection, we deploy Faster R-CNN \cite{fasterrcnn} by using ResNet50 as a backbone network on the PASCAL VOC dataset\cite{pascal2015}. During training, we apply LRPET to the backbone and train the model on PASCAL VOC 2007+2012 trainval dataset. We evaluate the performance with mean Average Precision (mAP) on the PASCAL VOC 2007 test dataset. The experimental results are shown in Table \ref{detection}. We can see that, the model pruned with LRPET reduce half of the  FLOPs with slight mAP loss. This demonstrates that our method generalizes well to objection detection.

\begin{table}[ht]
	\footnotesize
	\centering
	\caption{Results for Semantic Segmentation on PASCAL VOC 2012} 
	\label{segmentation}
	\centering
	\begin{tabular}{c|c|c|c}
		\hline
		Model & Method & FLOPs Reduction & mIOU \\
		\hline
		\multirow{2}*{PSPNet} & Baseline & 0\%   & 76.48   \\
		~ & LRPET & 30\% & 76.33 \\
		\hline
	\end{tabular}
\end{table}

For semantic segmentation, we apply LRPET to to the PSPNet \cite{pspnet}. We train the model on PASCAL VOC 2012 and Semantic Boundaries \cite{sbddata} training datasets, and evaluate the performance with mean Intersection over Union (mIOU) on the PASCAL VOC 2012 test dataset. The results are shown in in Table \ref{detection}. LRPET achieves considerable good performance. The mIOU of our method is slightly lower than the baseline while reducing 30\% of the FLOPs of the backbone network. This again verifies that LRPET has good generalization ability.

\subsection{Ablation Study}
Here we conduct ablation studies to investigate the contributions of each component in our proposed LRPET. ResNet-56 and VGG-16 on CIFAR-10 are taken into account. We train the networks from scratch. The pruning rate $P$ is set to  0.57 and 0.92 for ResNet-56 and VGG-16, respectively.

\begin{table}[ht]
	\footnotesize
    \centering
    \caption{Accuracy (\%) comparison on CIFAR-10 for energy transfer and BN rectification}
    \label{energy_com}
    \centering
    \begin{tabular}{c|c|c}
        \hline
        Method         & ResNet-56& VGG-16   \\
        \hline
        LRP &{92.68} &{92.77}\\
        LRP+ET & 92.94  & 92.82    \\
        LRP+BNR & 92.83  & 92.80   \\       
        {LRPET(LRP+ET+BNR)}& \textbf{92.99} & \textbf{92.93}\\
        \hline
    \end{tabular}
\end{table}

\subsubsection{Energy Transfer and BN Rectification} 
We study the effects of energy transfer and BN rectification by removing them. The experiment results are shown in Table \ref{energy_com}. Here ``LRP'' denotes applying low-rank projection at the end of each epoch without energy transfer or BN rectification. When using energy transfer without BN rectification, i.e., ``LRP+ET", the results are better than LRP. This verifies that energy transfer can ease the gradient energy reduction problem in Low-Rank Projection. Using BN rectification without energy transfer, i.e., ``LRP+BN rectification", also improves the results of Low-Rank Projection, which suggests that the incompatibility between BN and convolution for LRP is well solved by BN rectification. LRPET which adopts both strategies can further improve the performance. This shows that energy transfer and BN rectification have a collaborative effect on the final performance. Additionally, it's worth noting that most computation in LRPET come from the SVD in LRP. The incorporation of ET and BNR introduces only a few scalar product operations on the singular values and matrix multiplication with diagonal matrices. These additional operations are relatively negligible compared to the computational cost of SVD. Consequently, the performance gain achieved by LRPET over LRP is nearly free in terms of computational overhead.

\begin{figure}[t]
\centering
\subfloat[ResNet-56]{\includegraphics[width=1.72in]{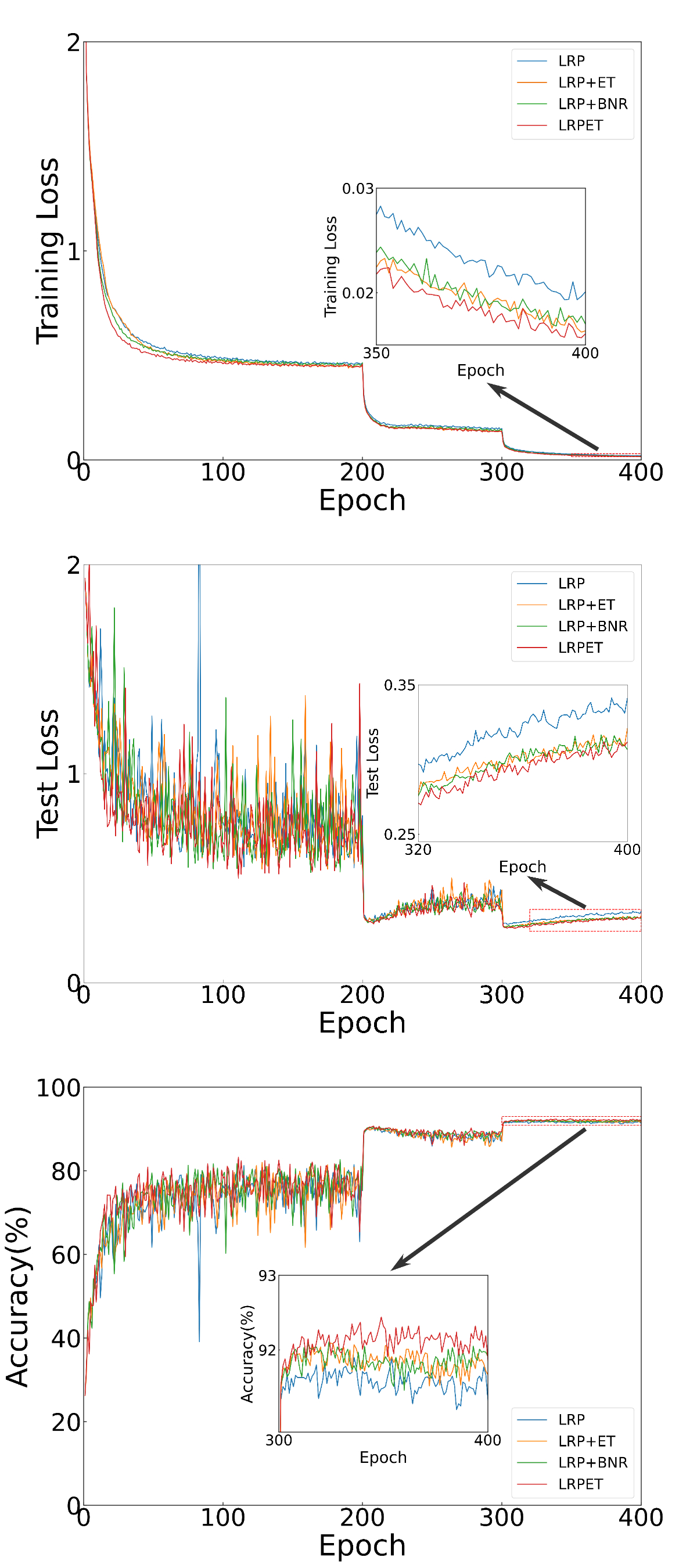}%
\label{res56_visual}}
\hfil
\subfloat[VGG-16]{\includegraphics[width=1.72in]{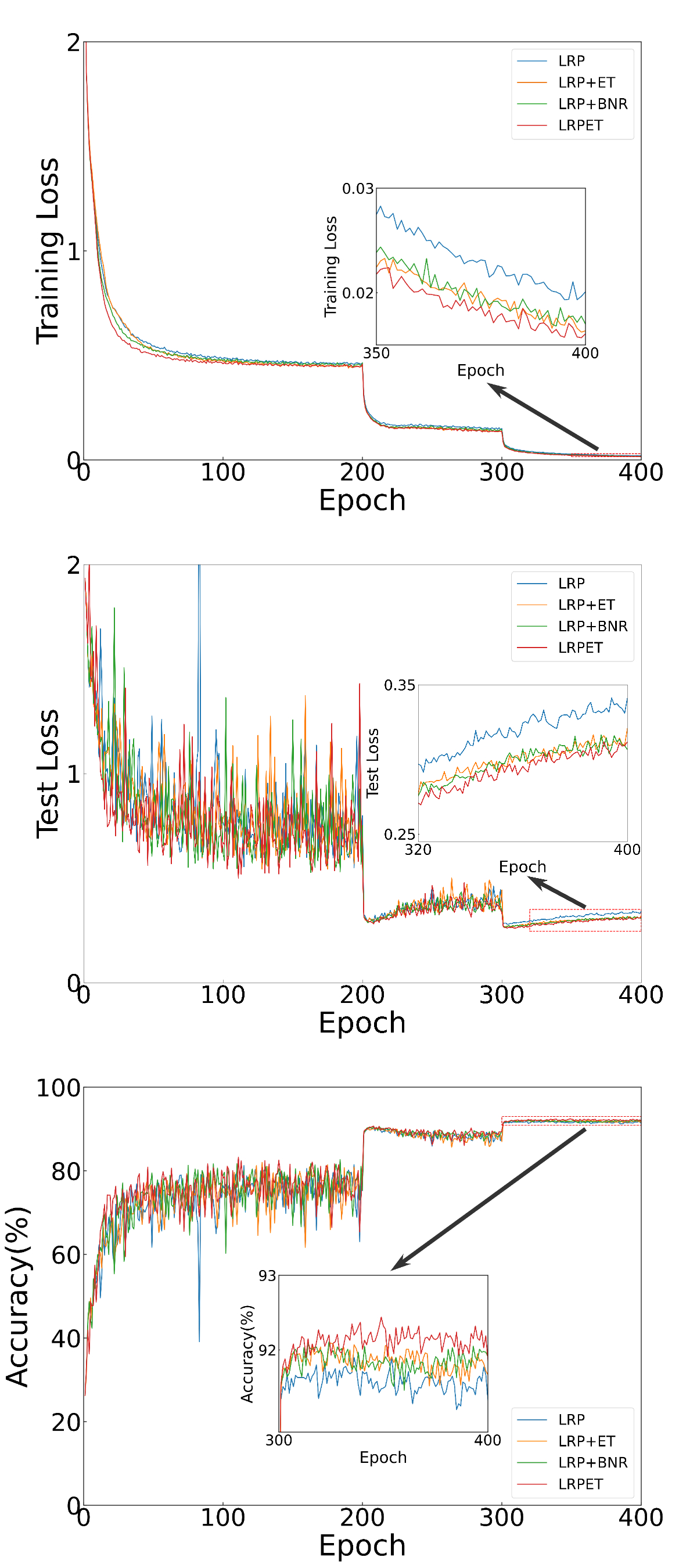}%
\label{VGG_visual}}
\caption{Learning Curves for LRP and its variants. The learning curves of training loss, test loss, and test accuracy for ResNet-56 and VGG-16 on CIFAR-10 are illustrated.}
\label{ablation_viusal}
\end{figure}

In Fig \ref{ablation_viusal}, we visualized the learning curves including training loss, test loss, and test accuracy of the experiments in Table \ref{energy_com}. Our optimization method is a variant of projected SGD (PSGD). The theoretical convergence analysis of PSGD has only been given for convex problems but the optimization of a deep neural network is highly nonconvex. Even for the simple SGD on a deep neural network, its convergence property is not thoroughly studied. However, PSGD works well in practice. We can see that, all the methods converge well in practice. In most experiments of LRPET, the relative errors between the original matrix and the product of the decomposed matrices for all the layers are smaller than 2\% and the accuracy of the decomposed network also maintains almost the same as the original network, which suggests that the optimized matrices are numerical low-rank.
Regarding the observed trends in the learning curves, the training loss shows slight but consistent improvement, especially noticeable in the more complex architecture of ResNet when compared to the plain VGG architecture. The apparent disparity in smoothness between training and test losses in our learning curves may arise from the differences in the  distributions of the training and test data.  We find that even in the absence of LRP and its variants, the test loss of baseline model also displays fluctuations.
This suggests that the observed phenomenon is inherent to the combination of SGD and variations in data distributions between training and test data. It is not specific to LRP and its variants but rather a characteristic of the optimization process when faced with differences in data distribution during training and testing phases. The fluctuations observed in the curves of test accuracy can be attributed to the unique characteristics of our approach, specifically the modification of weight matrices at the end of each epoch. This leads to more pronounced changes in weights compared to original SGD. However, as training progresses, the gaps between matrices before and after modification diminish, stabilizing the accuracy. This behavior is akin to phenomena observed in SVB (see Figure 2 in \cite{jia2017improving}). For enhanced clarity, the last stages of the figures were enlarged, revealing that LRP+ET, LRP+BNR, and LRPET exhibit noticeable improvements over the baseline LRP. 
Furthermore, the combination method LRPET consistently outperforms the single methods.

\subsubsection{Training Strategy} 
To show the superiority of our training strategy, we compare LRPET with four intuitive training strategies. The first one ``From Scratch'' is to decompose a random initialized network into low-rank form and train it from scratch. The second one ``AGD From Scratch'' is to decompose a random initialized network into low-rank form and train it from scratch, and then train it using the alternative gradient descent method. Specifically, each convolution layer is transformed into two convolution layers after low-rank decomposition, and the parameters of the two partial convolution layers are updated alternately during training. The third one ``Fine Tuning'' is directly decomposing a pre-trained baseline model and fine-tuning it. Here we fine-tune the network for 40 epochs with {a learning rate of 0.001.}
The fourth one ``Early Stop \& Ordinary Training'' refers to training the model with LRPET, stopping at the 200th epoch, and then continuing training the decomposed network to 400 epochs with the same learning rate as ordinary training. This strategy aims to investigate whether LRPET only provides a good initialization or works for the whole training process.
  
The results are shown in Table \ref{finetune}.
Training low-rank decomposition network from scratch performs the worst. This is because low-rank decomposition means lower model capacity and leads to a much deeper network, which makes optimization more difficult. Fine-tuning the fully trained model significantly improves the performance. However, it is still much worse than LRPET.
This suggests that a well-trained model in Euclidean space does not provide a good initialization for optimization with low-rank constraints.
An early stop of LRPET provides an initialization for a low-rank network but is not good enough. This is because the model is not close to the low-rank manifold enough at the stopped epoch. Thus, applying LRPET for the whole training process is needed.

\begin{table}[ht]
	\footnotesize
    \centering
    \caption{Accuracy (\%) comparison on CIFAR-10 with different training strategies } 
    \label{finetune}
    \centering
    \begin{tabular}{c|c|c}
        \hline
        Method       & ResNet-56  &VGG-16 \\
        \hline
        From Scratch & 62.62   & 57.59   \\
        AGD From Scratch & 59.01 & 52.27 \\
        Fine Tuning &  90.83 &  90.37 \\
        Early Stop \& Ordinary Training& 90.61& 91.41  \\
        {LRPET} & \textbf{92.99} & \textbf{92.93}\\
        \hline
    \end{tabular}
\end{table}

\subsubsection{Decompose Strategy} 
To further validate the soundness of our strategy of training on the original model before decomposing the network, we supplemented the ablation experiments with individual networks. The results are shown in Table \ref{scratch}. The network becomes narrow but deeper after Low-rank network compression, which makes it harder to train it. The strategy of decomposing before training leads to terribly bad results. So training on the original model before decomposing is useful.

\begin{table}[ht]
	\footnotesize
    \centering
    \caption{Accuracy (\%) comparison on CIFAR-10 with different decompose strategies } 
    \label{scratch}
    \centering
    \begin{tabular}{l|c|c}
        \hline
        Method       & From Scratch  & LRPET \\
        \hline
        Resnet-56(${P}$=0.57) & 62.62   & 92.99   \\
        Resnet-110(${P}$=0.65) & 71.69 & 93.61 \\
        VGG-16(${P}$=0.92) &  57.59 &  92.93 \\
        GoogleNet(${P}$=0.60) & 80.85 & 95.25  \\
        FBNet(${P}$=0.40) & 86.15 & 94.66 \\
        \hline
    \end{tabular}
\end{table}

\subsubsection{Amplification coefficient {$\alpha$}} 
\begin{figure}[t]
\centering
\subfloat[ResNet-56]{\includegraphics[width=1.72in]{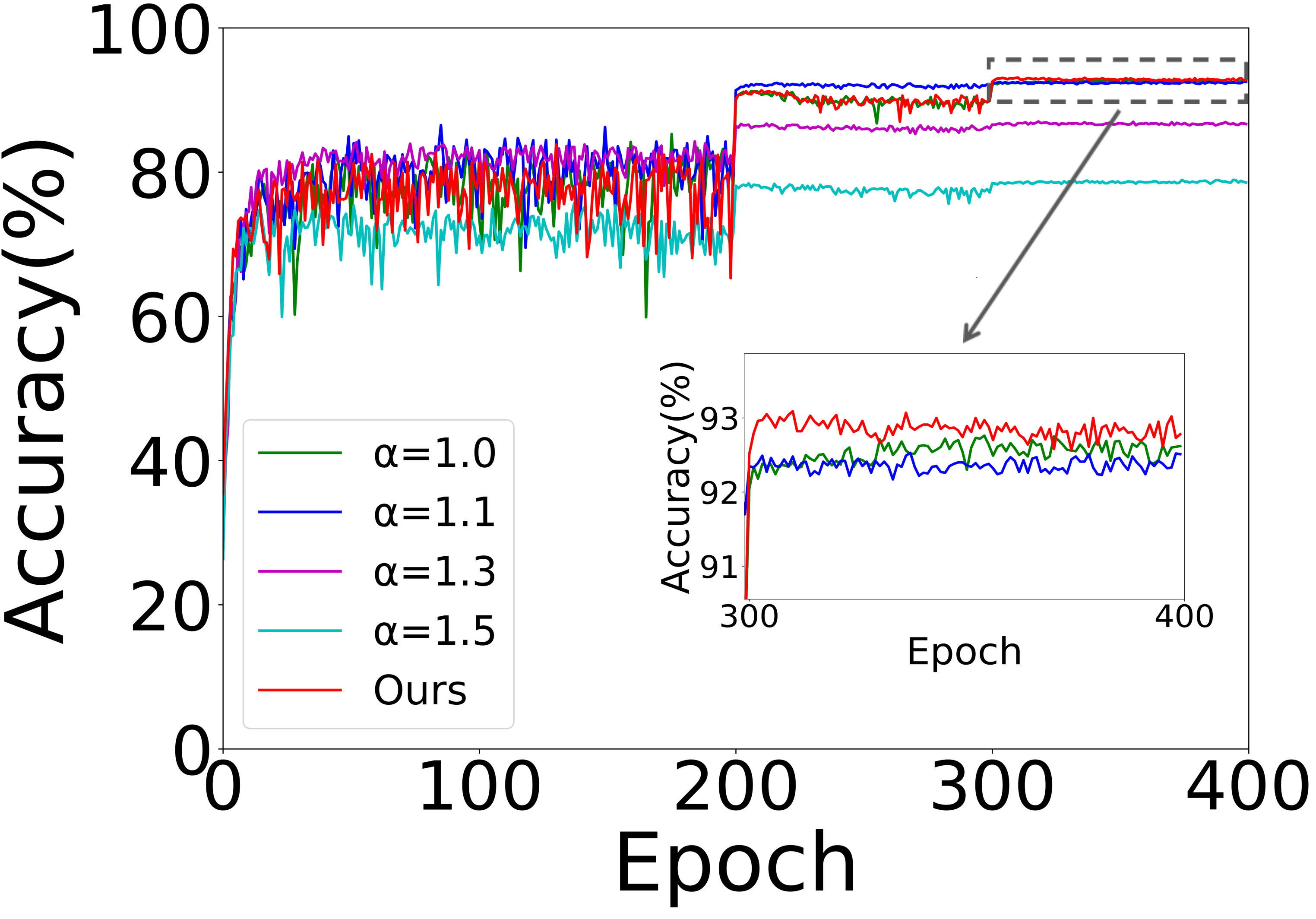}%
\label{res56add_case}}
\hfil
\subfloat[VGG-16]{\includegraphics[width=1.72in]{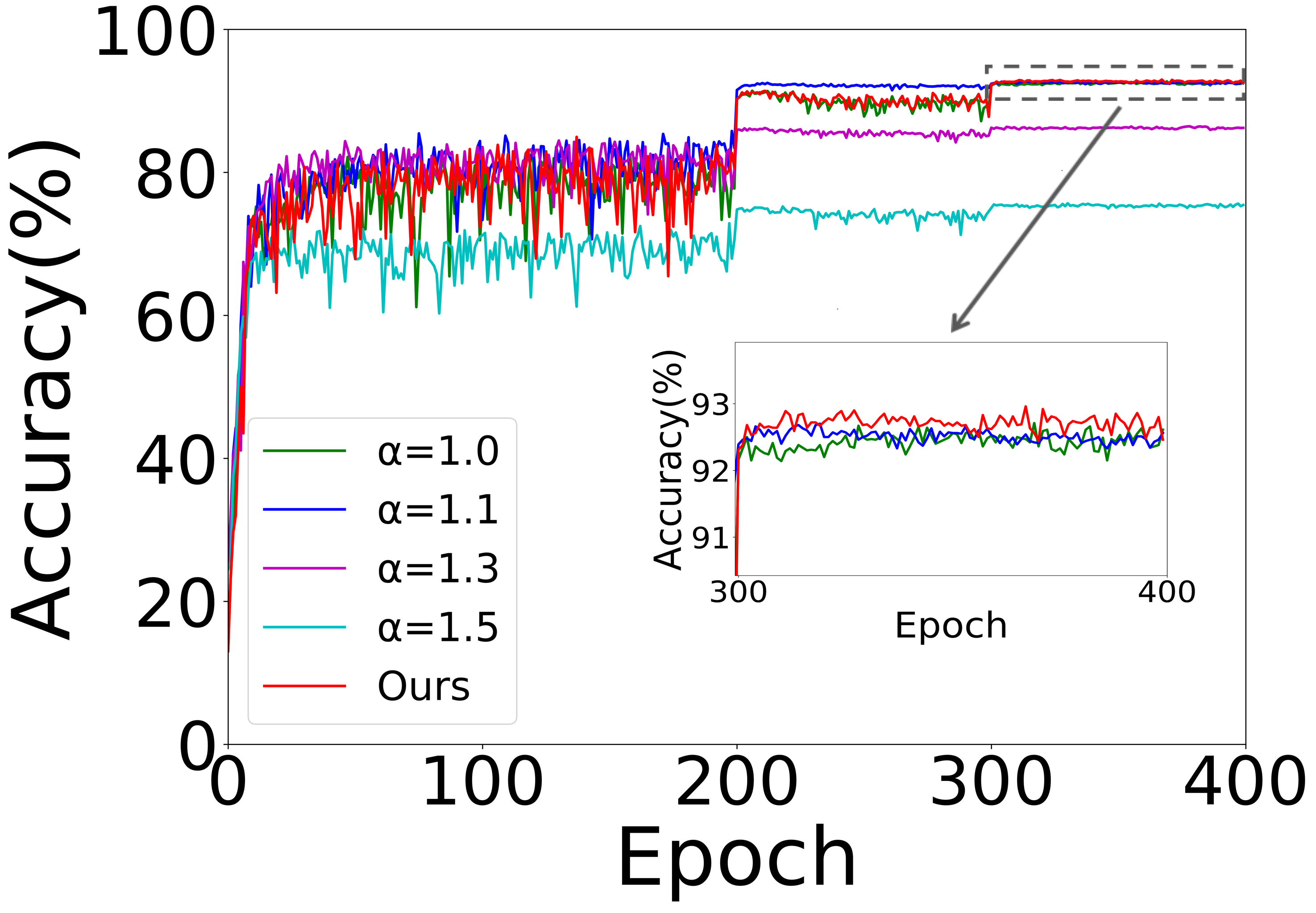}%
\label{VGGadd_case}}
\caption{Test accuracy curves of different fixed $\alpha$ and our adaptive amplification coefficient $\alpha$ setting for ResNet-56 and VGG-16 on CIFAR-10 . }
\label{coefficient}
\end{figure}

We transfer energy with an adaptive amplification coefficient $\alpha$ to keep the energy. To verify the effectiveness of such a strategy, we compared the performance of the network using different fixed values of $\alpha$ and our adaptive strategy. Fig. \ref{coefficient} illustrates the test accuracy curves of different $\alpha$ settings for ResNet-56 and VGG-16 on CIFAR-10.
When $\alpha$ is small, the accuracy can be improved, which shows that the reduction of gradient energy will degrade the performance and can be recovered by energy compensation. When $\alpha$ is too large, the weight will deviate from the optimal solution and the result is poorer than without energy compensation. In contrast, the adaptive amplification coefficient $\alpha$ has better performance than all fixed values of $\alpha$, which demonstrates that keeping the energy of the weight matrix the same is effective for training low-rank networks. 

\subsection{Combination with Other Methods}
To explore the scalability of our approach, we try to combine LRPET with state-of-the-art compression alternatives including hashing and quantization methods.

We combine our approach with SMH. We first train a low-rank compact network via LRPET, and then train the low-rank network with SMH. The results of ResNet-56 and VGG-16 on CIFAR-10 are shown in Table \ref{hash}. We can see that LRPET is compatible with the hashing method and can further improve the performance of SMH.

\begin{table}[ht]
	\footnotesize
	\centering
	\caption{Hashing Integration on CIFAR-10}
	\label{hash}
	\scalebox{1}{
		\centering
		\begin{tabular}{l|c|c|c}
			\hline
			Model      & Method &  Parmas & Top-1 Acc. (\%) \\
			\hline
			\multirow{4}*{ResNet-56} & SMH & 166.39k & 89.67  \\
			~ & LRPET+SMH  & 162.68k & 90.14  \\
            \cdashline{2-4}
			~ & SMH  & 243.84k & 90.24  \\
			~ & LRPET+SMH  & 241.96k & 90.60  \\
            \hline
            \multirow{4}*{VGG-16} & SMH & 1.48M & 90.14  \\
			~ & LRPET+SMH  & 1.23M & 91.13  \\
            \cdashline{2-4}
			~ & SMH  & 1.64M & 90.20  \\
			~ & LRPET+SMH  & 1.56M & 91.03  \\
			\hline
		\end{tabular}
	}
\end{table}

We combine LSQ and LPRET to obtain higher accuracy with lower bit operations. Specifically, we train a low-rank network with LRPET and then quantize it with LSQ. The results of ResNet-56 and VGG-16 on CIFAR-10 are shown in Table \ref{lsq}. Compared to quantizing the original network directly, The combination method achieves higher accuracy with a lower number of bit operations.

\begin{table}[ht]
	\footnotesize
	\centering
	\caption{Quantization Integration on CIFAR-10}
	\label{lsq}
	\scalebox{1}{
		\centering
		\begin{tabular}{l|c|c|c|c}
			\hline
			Model      & Method & W/A &  BiTs & Top-1 Acc. (\%) \\
			\hline
			\multirow{2}*{ResNet-56} & LSQ  & 5/5 & 3.12B & 93.80  \\
			~ & LRPET+LSQ  & 5/6 & 2.87B & 93.90  \\
            \hline
            \multirow{2}*{VGG-16} & LSQ  & 4/4 & 5.01B & 94.03  \\
            ~ & LRPET+LSQ & 4/5 & 4.95B & 94.10  \\
			\hline
		\end{tabular}
	}
\end{table}

''W/A'' represents bit-widths of of weights and activations, and ''BiTs'' represents the bit operands of model. Specifically, assume \textbf{x} is a M bit-width integer and \textbf{y} is a k bit-width integer, the computation complexity about the multiplication of \textbf{x} and \textbf{y} is \textit{O(MK)}, directly proportional to bitwidth of \textbf{x} and \textbf{y}\cite{zhou2016dorefa}.

\subsection{Gradient Norm Analysis}
Here we empirically compute the gradient norm to verify the theoretical analysis in Section \ref{ET}. The experiments of ResNet-56 are conducted on CIFAR-10 by applying LRP and LRPET, respectively. We compute the gradient norms of the weights of the first and second convolution layers of ResNet-56 before and after applying LRP and LRPET, respectively, and average them among all epochs. The average gradient norms and relative decreases are shown in Table \ref{gradient norms}. We can see that LRP  decreases the gradient norm and LRPET can ease this trend.

\begin{table}[ht]
	\footnotesize
	\centering
	\caption{Gradient Norms Comparison on CIFAR-10}
	\label{gradient norms}
	\scalebox{1}{
		\centering
		\begin{tabular}{l|c|c|c|c}
			\hline
			Layer      & Method & Before&  After & Relative Decrease (\%) \\
			\hline
			\multirow{2}*{Conv1} & LRP  & 0.2531 & 0.2473 & 2.29  \\
			~ & LRPET  & 0.2577 & 0.2537 & 1.55  \\
			\hline
			\multirow{2}*{Conv2} & LRP  & 0.1616 & 0.1586 &  1.86 \\
			~ & LRPET & 0.1875 & 0.1857& 0.96  \\
			\hline
		\end{tabular}
	}
\end{table}

\subsection{Training Time Consumption}\label{time_consumption}
Although the computation complexity of SVD is high, SVD is only used at the end of each epoch or a certain number of iterations in LRPET and the training time increment is small. TRP \cite{DBLP:conf/ijcai/XuL0WWQCLX20} also utilizes SVD for training, but it is very time-consuming since it needs SVD for every iteration. In Table  \ref{using_time}, we show the average training time of one epoch of LRPET and TRP for ResNet-56, ResNet-110, and VGG-16 on CIFAR-10 for comparison.
Compared to the original SGD, the relative additional training time of LRPET is small, while TRP takes multiple times of training time of LRPET. When nuclear norm regularization is included, the training time of TPR will further increase. For example, in training ResNet-56 on CIFAR-10, the relative additional training time of LRPET compared to original SGD is only +3.2$\%$, while TRP is +371$\%$, which verifies that our LRPET is a simple and efficient method.

\begin{table}[ht]
	\footnotesize
	\centering
	\caption{Training time (seconds)  on CIFAR-10}
	\label{using_time}
	\scalebox{1}{
		\centering
		\begin{tabular}{l|c|c|c}
			\hline
			Method         & ResNet-56 & ResNet-110 & VGG-16   \\
			\hline
			Baseline & 25.0   &48.2 & 17.2 \\
			\hdashline
			LRPET & 25.8 (+3.2\%) &48.6 (+0.8\%) & 18.8 (+9.3\%)    \\
			TRP \cite{DBLP:conf/ijcai/XuL0WWQCLX20} & 117.8 (+371\%) &229.3 (+376\%) & 171.8 (+899\%)\\
			TRP+Nu \cite{DBLP:conf/ijcai/XuL0WWQCLX20} & 139.2 (+457\%) & 241.2 (+400\%) & 175.8 (+922\%)\\
			\hline
		\end{tabular}
	}
\end{table}

\section{Conclusion}
\label{sec:conclu}
Low-rankness is a popular property for compression in traditional learning, but it is not so popular in compressing deep neural networks. This is because a neural network with low-rank constraint is hard to train. Compared to other network compression methods, low-rank-based methods seem less competitive. In this paper, we propose a new training method that efficiently trains low-rank constraint network with considerable good results. We reveal that the potential effectiveness of low-rankness is neglected in network compression. LRPET not only outperforms the other low-rank compression method but also some state-of-the-art pruning methods.

\bibliographystyle{IEEEtran}
\bibliography{egbib}

\begin{IEEEbiography}[{\includegraphics[width=1in,height=1.25in]{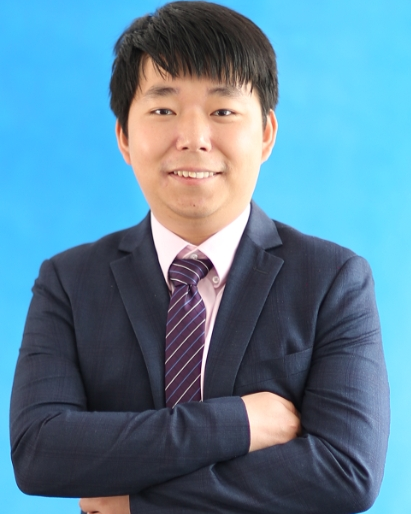}}]{Kailing Guo}(Member, IEEE)
	received the Ph.D. degree from the South China University of Technology, Guangzhou, China. He is currently an Associate Professor with the School of Electronic and Information Engineering, South China University of Technology. His research interests include low-rank and sparse learning, deep learning optimization and model compression, multimodal human data processing.
\end{IEEEbiography}

\vspace{0.1in}

\begin{IEEEbiography}[{\includegraphics[width=1in,height=1.25in]{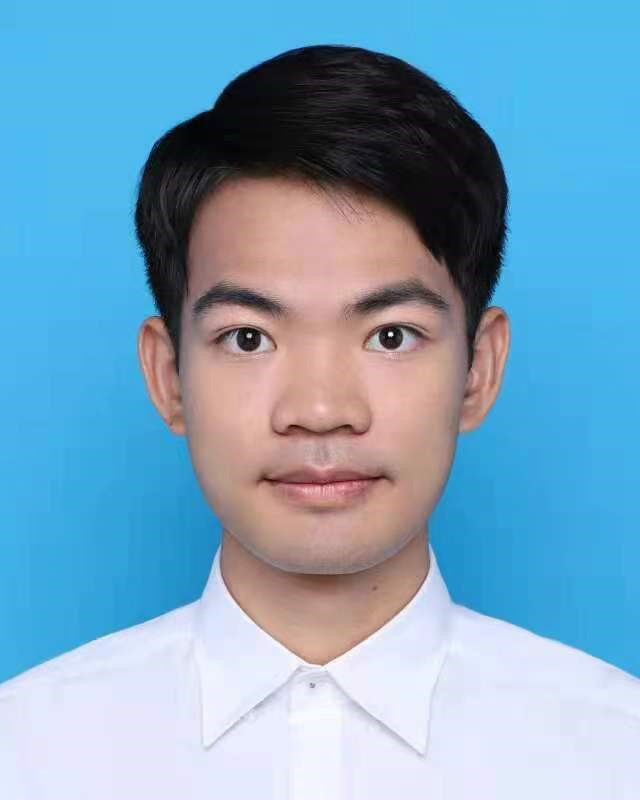}}]{Zhenquan Lin} received the B.E. degree from South China Agricultural University in 2020 and the M.E. degree in electronic information from South China University of Technology in 2023. His research focuses on neural network compression and computer vision.
\end{IEEEbiography}

\vspace{0.1in}

\begin{IEEEbiography}[{\includegraphics[width=1in,height=1.25in]{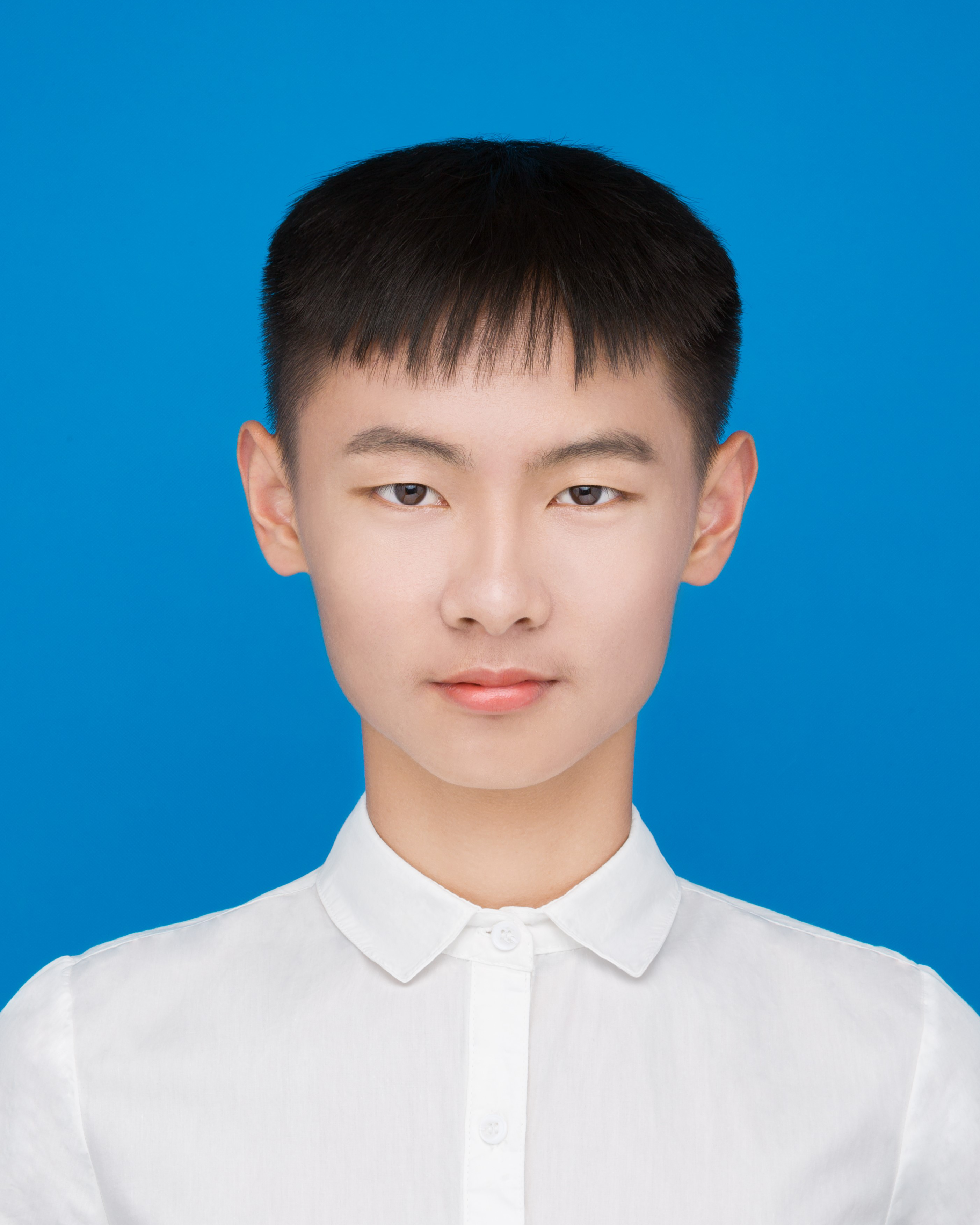}}]{Canyang Chen} received the B.E. degree from South China University of Technology in 2023 and is now pursuing the M.E. degree in electronic information from South China University of Technology. His research focuses on neural network compression and computer vision.
\end{IEEEbiography}

\vspace{0.1in}

\begin{IEEEbiography}[{\includegraphics[width=1in,height=1.25in]{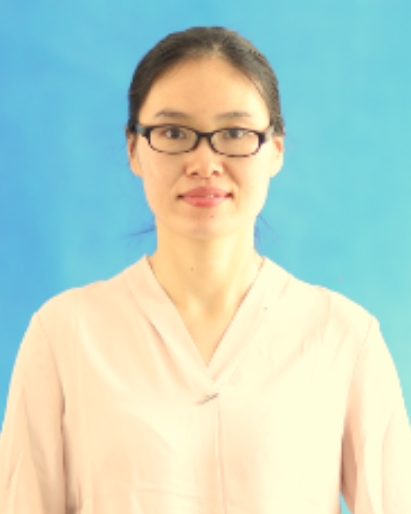}}]{Xiaofen Xing}(Member, IEEE) received the B.S.,
	M.S., and Ph.D. degrees from the South China University of Technology, Guangzhou, China, in 2001,
	2004, and 2013, respectively. Since 2017, she has
	been an Associate Professor with the School of Electronic and Information Engineering, South China
	University of Technology. Her main research interests
	include speech emotion analysis, image/video processing, and human computer interaction.
\end{IEEEbiography}

\vspace{0.1in}

\begin{IEEEbiography}[{\includegraphics[width=1in,height=1.25in]{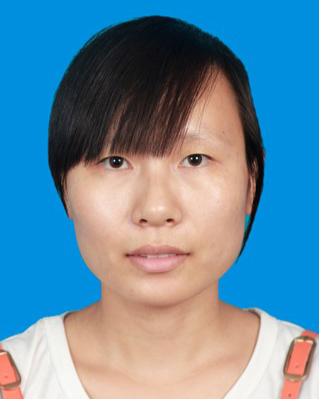}}]{Fang Liu} received the Ph.D. degree from the South China University of Technology, Guangzhou, China. She is  currently a lecturer with the School of Internet Finance and Information Engineering, Guangdong University of Finance. Her research interests include transfer learning, privileged information, and action recognition.
\end{IEEEbiography}

\vspace{0.1in}

\begin{IEEEbiography}[{\includegraphics[width=1in,height=1.25in]{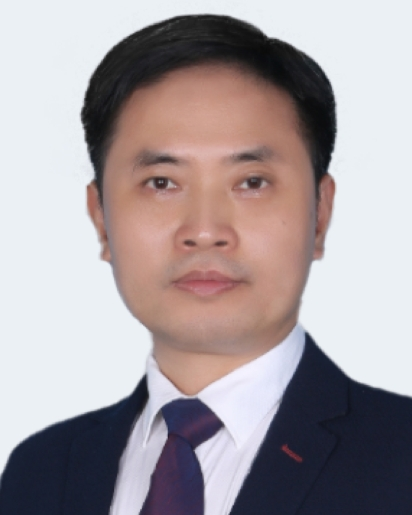}}]{Xiangmin Xu}
	(Senior Member, IEEE) received the
	Ph.D. degree from the South China University of
	Technology, Guangzhou, China. He is currently a Full
	Professor with the School of Electronic and Information Engineering and the School of Future Technology, South China University of Technology. 
	His recent research focuses on image/video processing, human–computer interaction, computer vision, and machine learning.
	
\end{IEEEbiography}

\end{document}